\documentclass[sigconf,anonymous=false,nonacm=true]{acmart}

\usepackage[utf8]{inputenc}
\usepackage{amsmath,mathtools} % amssymb messes things with Bbbk
\usepackage[inline]{enumitem}
\usepackage[detect-weight=true, binary-units=true]{siunitx}
\usepackage{url,hyperref,cleveref}
\crefname{algocf}{alg.}{algs.}
\Crefname{algocf}{Algorithm}{Algorithms}
\Crefname{equation}{Eq.}{Eqs.}
\Crefname{figure}{Fig.}{Figs.}
\Crefname{tabular}{Tab.}{Tabs.}
\Crefname{section}{Sec.}{Secs.}
\usepackage[linesnumbered]{algorithm2e}
\crefname{algocf}{algorithm}{algorithms}
\Crefname{algocf}{Algorithm}{Algorithms}
\usepackage{color,colortbl,etoolbox,xcolor} %http://tex.stackexchange.com/a/2768/22613
\robustify\bfseries
\usepackage{multirow,booktabs}
\usepackage{subcaption}
\usepackage{csquotes} %http://tex.stackexchange.com/a/294634
\usepackage[normalem]{ulem} %http://tex.stackexchange.com/a/23712
\renewcommand\vec{\boldsymbol}

\usepackage{comment}

\newcommand{\name}{ML-PIE}

\DeclareRobustCommand{\pie}{%
  \begingroup\normalfont
  \includegraphics[height=1.5\fontcharht\font`\B]{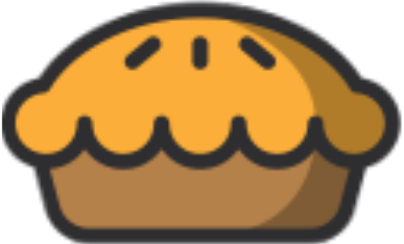}%
  \endgroup
}

\acmDOI{10.1145/nnnnnnn.nnnnnnn} % To be updated after completing copyright process
\acmISBN{978-x-xxxx-xxxx-x/YY/MM} % To be updated after completing copyright process
\acmConference[GECCO '21]{The Genetic and Evolutionary Computation Conference 2021}{July 10--14, 2021}{Lille, France}
\acmYear{2021}
\copyrightyear{2021}

\begin{document}

\title[Model Learning with \pie]
{Model Learning with Personalized Interpretability Estimation\\
(ML-\includegraphics[width=0.05\linewidth]{pics/thepie.pdf})}

% arlenative titles:
% Interpretable artificial intelligence at a personalized level: an active learning approach with neural networks and genetic programming

\author{Marco Virgolin}
\email{marco.virgolin@chalmers.se}
\affiliation{%
  \institution{Chalmers University of Technology}
  \city{Gothenburg}
  \country{Sweden}
}

\author{Andrea De Lorenzo}
\email{andrea.delorenzo@units.it}
\affiliation{%
  \institution{University of Trieste}
  \city{Trieste}
  \country{Italy}
}

\author{Francesca Randone}
\email{francesca.randone@imtlucca.it}
\affiliation{%
  \institution{IMT Lucca}
  \city{Lucca}
  \country{Italy}
}

\author{Eric Medvet}
\email{eric.medvet@units.it}
\affiliation{%
  \institution{University of Trieste}
  \city{Trieste}
  \country{Italy}
}

\author{Mattias Wahde}
\email{mattias.wahde@chalmers.se}
\affiliation{%
  \institution{Chalmers University of Technology}
  \city{Gothenburg}
  \country{Sweden}
}

\renewcommand{\shortauthors}{Virgolin et al.}

\begin{abstract}
High-stakes applications require AI-generated models to be interpretable.
Current algorithms for the synthesis of \emph{potentially} interpretable models rely on objectives or regularization terms that represent interpretability only coarsely (e.g., model size) and are not designed for a specific user.
Yet, interpretability is intrinsically subjective.
In this paper, we propose an approach for the synthesis of models that are tailored to the user by enabling the user to steer the model synthesis process according to her or his preferences.
We use a bi-objective evolutionary algorithm to synthesize models with trade-offs between accuracy and a user-specific notion of interpretability. 
The latter is estimated by a neural network that is trained concurrently to the evolution using the feedback of the user, which is collected using uncertainty-based active learning.
To maximize usability, the user is only asked to tell, given two models at the time, which one is less complex. 
With experiments on two real-world datasets involving 61 participants, we find that our approach is capable of learning estimations of interpretability that can be very different for different users. 
Moreover, the users tend to prefer models found using the proposed approach over models found using non-personalized interpretability indices.
\end{abstract}

\begin{teaserfigure}
    \includegraphics[width=\textwidth]{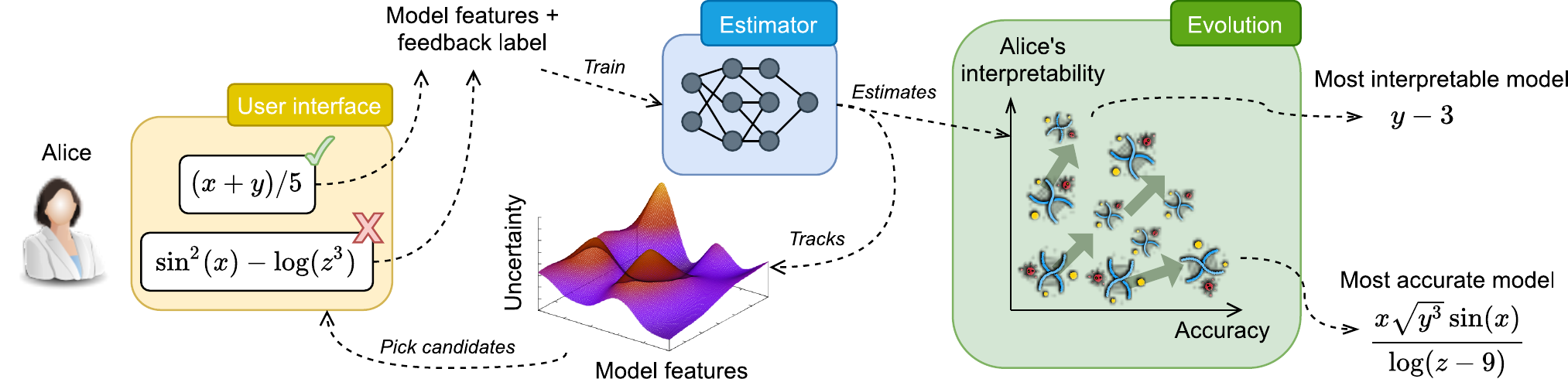}
    \caption{
        Schematic view of the proposed approach, \name{}. 
        In the implementation proposed in this paper,
        the user provides feedback on models that are being discovered by an evolutionary algorithm.
        This feedback is used to train an estimator which, in turn, shapes one of the objective functions used by the evolution. 
        Ultimately, this steers the evolution towards discovering models that are interpretable according to the specific user.
        To minimize the amount of feedback needed, \name{} keeps track of which models cause the estimator to be most uncertain, and submits these models for user assessment. 
    }
    \label{fig:teaser}
\end{teaserfigure}




\maketitle

\section{Introduction}
World-wide policies concerning the fair and responsible use of artificial intelligence (AI) increasingly demand AI-made decisions to be explainable~\cite{jobin2019global}.
To answer this call, the field of \emph{eXplainable AI} (XAI) studies methods to provide explanations of the decisions taken by unintelligible, \emph{black box} models, such as deep neural networks~\cite{adadi2018peeking}.
However, explanation methods can be limited and should be used with care.
For example, these methods may produce explanations that are only valid for a narrow neighborhood of the input data; mislead about feature relevance when importance scores are scattered across correlated features; and provide seemingly sensible explanations even when the black box is blatantly wrong~\cite{molnar2020interpretable,rudin2019stop}.
For these reasons, XAI also studies how to develop methods that can produce \emph{interpretable AI}, i.e., models which can be inspected by humans as to provide a complete picture of their workings (including, e.g., edge cases). 
When capable of good performance, interpretable models should be preferred over black box ones~\cite{molnar2020pitfalls,rudin2019stop}.

Evolutionary computation provides effective methods to seek interpretable AI models, and genetic programming (GP)~\cite{koza1992genetic} is a prime candidate in this sense.
GP evolves a population of programs, such as AI models (or, more specifically, machine learning models), made of arbitrary primitive instructions.
If these instructions are chosen to be procedures that perform high-level computations and have a clear meaning, GP has the \emph{potential} to discover models that are, themselves, capable of complex functional behavior while also being interpretable.
Nevertheless, GP potential to discover interpretable models is likely to remain unexpressed if left to chance.
In other words, it is desirable to encapsulate the concept of interpretability into a search objective, so that the search process of GP can be steered towards most promising models.

The design of an objective that represents interpretability is an open problem.
To begin with, there exists no clear-cut definition of what interpretability is~\cite{lipton2018mythos,benk2020explaining}.
Whether a person finds a model interpretable depends on that person's background and, further adding to the complexity, the sensitiveness of the application at hand plays a role in deciding what \emph{degree} of interpretability may be sought~\cite{freitas2014comprehensible,hatherley2020limits}.
Paradoxically, to seek clear, interpretable models, we need an objective that cannot be clearly defined.

In this paper, we tackle the problem of capturing the subjective notion of interpretability using GP and \emph{active learning}.
Instead of attempting to decide beforehand what the user deems interpretable, we train a personalized estimator of interpretability (in the form of a neural network) from feedback that is given by the user during the model synthesis process (in our case, GP).
The user provides feedback by telling, given two models at the time, which one is more interpretable; this feedback refines the estimator.
Concurrently, the estimator steers the model synthesis process (by implementing an objective of GP).
To train the estimator sufficiently well from limited feedback, we keep track of what models cause the estimator to be most uncertain and submit those models for human assessment.
\Cref{fig:teaser} shows the overall approach, which we call \emph{Model Learning with Personalized Interpretability Estimation} (\name). 

We describe how we realize ML-PIE and show that it is capable of tackling the problem of discovering models with personalized interpretability,
potentially paving the way for a new generation of personalized XAI methods.

%We realize ML-PIE as a web application and share the weblink with Engineering students enrolled at the University of Trieste, Italy, and Chalmers University of Technology.
%By analyzing data regarding the progress and outcome of our approach, we find that ML-PIE is indeed capable of tailoring the search process to the user, hence showing that it is possible to design human-centric algorithms for the discovery of models with personalized interpretability.

\section{Related work}\label{sec:related-work}
Linear models, decision tables, decision trees, and other models that consist of high-level rules, are considered to have good chances of being  interpretable~\cite{huysmans2011empirical,guidotti2018survey}.
For linear models, promoting interpretability essentially corresponds to reducing the number of features~\cite{tibshirani1996regression,zou2005regularization,ustun2016supersparse,poursabzi2018manipulating}. 
For decision trees and decision rules, besides reducing the number of features, approaches exist to restrict model size, prune unnecessary parts~\cite{breslow1997simplifying,izza2020explaining}, aggregate local models in a hierarchy~\cite{setzu2021glocalx}, or promote a trade-off between accuracy and complexity by means of loss functions~\cite{lakkaraju2016interpretable, su2015interpretable} or prior distributions~\cite{wang2017bayesian, letham2015interpretable, wang2015falling}. 
Regarding GP (and close relatives like grammatical evolution), perhaps the most simple and popular strategy to favor interpretability is to restrain the number of model components~\cite{ekart2001selection,virgolin2020improving,lensen2021mining}, sometimes in elaborate ways or particular settings~\cite{smits2005pareto,cano2013interpretable,virgolin2020explaining,lensen2020genetic,mota2021towards}.
Another strategy consists of penalizing models according to a weighted sum of the components that they include, after having pre-determined a weighing scheme~\cite{hein2018interpretable,medvet2015evolutionary}.
Alternatively, information from model approximators such as polynomials can be used to determine the level of interpretability~\cite{vladislavleva2008order}.

Recently, an estimator of human-interpretability for symbolic models expressed as formulae was machine-learned from human feedback~\cite{virgolin2020learning}.
A survey was used to make users simulate the calculations of (random) formulae (an implementation of the XAI concept of \emph{simulatability}~\cite{lipton2018mythos}), and to identify the behavior of the formula when part of it would vary in some interval (an implementation of the XAI concept of \emph{decomposability}~\cite{lipton2018mythos}).
Data gathered from \num{334} users was then used to fit a linear estimator comprised of four features extracted from the formulae, as, e.g., the subsequent composition of non-arithmetic operations.
The estimator of interpretability was finally incorporated into a bi-objective GP, to evaluate the interpretability of evolving models. 
This estimator has also been used in another recent work~\cite{custode2020evolutionary}.

In this paper we build upon~\cite{virgolin2020learning} and extend it in three ways:
\begin{enumerate*}[label=(\arabic*)]
    \item we \emph{concurrently} learn a model of intepretability that is specific to the user;
    \item we use a more complex, non-linear estimator instead of a linear one; and
    \item we exploit uncertainty estimation to require a small amount of feedback to train the estimator.
\end{enumerate*}

Several works have explored including humans in the training loop of AI system synthesis for different applications, e.g.,~\cite{secretan2011picbreeder,christiano2017deep,mahoor2017morphology}. 
Very recently, in~\cite{mota2021towards} it is hypothesized that querying humans during an (evolutionary) model synthesis process may help discovering interpretable models: this is what ML-PIE realizes.
Furthermore, since ML-PIE actively seeks feedback about the models that make the estimator most uncertain, our work is a form of \emph{active learning}. 
Broadly speaking, active learning entails techniques to automatically identify what few data items should be labeled, when labeling is expensive (e.g., when a user is involved)~\cite{settles2009active,yang2016active}.
Here we use a relatively simple strategy: we submit for user assessment those models that cause the estimator to be most uncertain.
There are a few works that used active learning together with GP, similarly to us, e.g.,~\cite{de2010active,isele2013active,bartoli2017active}.
An interesting difference that these works have with ours is that they use the labels acquired with active learning within the (or an, if multiple) objective function of GP \emph{directly}, whereas we use these labels to \emph{update} an objective function, namely the estimator of the interpretability of the user.

\section{\name{} overview}

\paragraph{Problem statement}
We consider the case of a user interested in obtaining a \emph{model} by means of some model synthesis process.
For the sake of simplicity, we assume the latter to be a supervised learning algorithm (but needs not be).
We assume that some data are available to synthesize (and validate) the model, e.g., $\{(\vec{x}_i, y_i)\}^n_{i=1}$ with $\vec{x}_i \in \mathbb{R}^d$ a vector of feature values and $y_i \in \mathbb{R}$ a label value, and that the user desires the model to be
\begin{enumerate*}[label=(\alph*)]
    \item accurate in terms of estimating a $y_i$ when given an $\vec{x}_i$ as input, and
    \item interpretable.
\end{enumerate*}
We do not enforce a precise definition of interpretabilty.
Instead, we only assume that the user, when asked to compare two models, is able to choose the most interpretable one.

\paragraph{Use case}
From the point of view of the user, \name{} works as follows: once the user has provided the data to the system, the model synthesis process starts.
In \name{}, the model synthesis process takes into account both the accuracy and the interpretability of the model being synthesized: the accuracy is measured with an appropriate index (e.g., accuracy for classification or negative mean squared error for regression), the interpretability is estimated according to a dedicated \emph{estimator} that is updated during the process---in fact, both the model under synthesis and the estimator of interpretability can be called ``model'' or ``estimator'', and both are ``synthesized'', ``trained'', or ``learned'' according to some machine learning algorithm; for the sake of clarity, we choose to use a different terminology to distinguish them.

\begin{figure}
    \centering
    \setlength{\tabcolsep}{0pt}
    \renewcommand{\arraystretch}{0}
    \begin{tabular}{cc}
        \includegraphics[width=\linewidth]{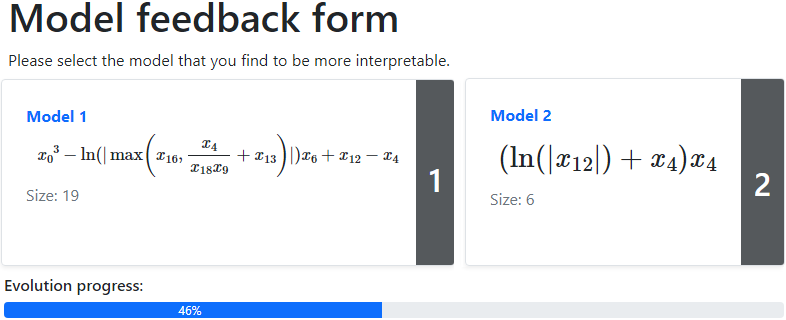} & 
    \end{tabular}
    \caption{
        Screenshot of the interface used to collect the feedback of the user. 
        The user clicks (or taps) on the model he or she deems more interpretable.
        Two new models are shown next.
        On the bottom, the progress bar displays the progress of the model synthesis process.
    }
    \label{fig:feedback-form}
\end{figure}

While synthesizing the model, the system shows the user, through a suitable user interface, a \emph{query}: given a pair of models, the user is asked to tell which one is the most interpretable according to her or his subjective judgment.
As soon as the user answers the query (by a single click or tap), \name{} shows a new query; internally, it also uses the feedback to update the interpretability estimator.
Query-answer interactions continue until the model synthesis process stops because a termination criterion is met.
We remark that query-answer interactions and the model synthesis process proceed concurrently and asynchronously (i.e., we do not halt the model synthesis process at any point in time, see~\Cref{sec:discussion}).
The interpretability estimator acts as point of contact between the two, as it is updated right after a user's feedback and is used by the model synthesis process to assess model quality.
\Cref{fig:feedback-form} shows the user interface we used: the user answers the query by simply clicking on the preferred model (in our case, models are formulae).

\paragraph{Design goals}
While designing \name{} we were driven by two goals.
First, we wanted to keep low the annotation effort: we do not ask the users to provide detailed descriptions of what they mean by interpretability, neither do we ask for detailed, informative feedback about the models.
Rather, we just ask to compare pairs of models and answer by performing an immediate action (single click or tap).
Second, we aimed at obtaining a system that can learn models that are interpretable for a specific user on a specific problem, i.e., we wanted a \emph{personalized} notion of interpretability.
To achieve this goal, we decided to update the interpretability estimator concurrently to the model synthesis process.
Note that, this way, the models undergoing user assessment necessarily come from the distribution of models under synthesis.
In contrast, performing an off-line annotation approach prior to model synthesis would require to decide a sample of possible models beforehand, some of which may turn out to be unrepresentative of those under synthesis, ultimately wasting part of the user's effort.
Even if ML-PIE updates the estimator while the model synthesis process takes place, we highlight that ML-PIE does not have to start from scratch: a partly pre-trained estimator can be used, as well as one that was previously built for other users, or a similar domain (e.g., healthcare).

\paragraph{Generality and requirements}
In principle, our approach is agnostic with respect to the algorithm employed for the model synthesis process and the nature of the models being synthesized, provided that:
\begin{enumerate*}[label=(\alph*)]
    \item the former is able to pursue accuracy and interpretability at the same time, either separately with a multi-objective algorithm or jointly with a single-objective one (e.g., to maximize an overall index of \emph{trust}, we expand on this in~\Cref{sec:discussion}); 
    \item the latter allows for a visual representation on a user interface (e.g., formulae or decision trees).
\end{enumerate*}
A further, more practical requirement, is that the speed of the model synthesis process is compatible with allowing the user to provide a sufficient amount of feedback.
Our approach is, in principle, also agnostic with respect to the algorithm employed for updating the estimator and the nature of the estimator, provided that these allow for active learning.
This last requirement can be met if, e.g., the estimator can produce a measure of uncertainty or confidence when estimating the interpretability of the models~\cite{de2010active}.

In this paper, we instantiate ML-PIE for symbolic regression, i.e., our models are symbolic functions expressed as formulae. 
We use multi-objective GP as model synthesis process.
For the estimator, we use a neural network and train it with stochastic gradient descent; to make the network produce predictions with uncertainty, we use dropout at prediction time.
In the next section, we describe in detail these key internals of \name{}.

\section{\name{} internals}
We choose to use GP in a multi-objective setting because its outcome is a \emph{collection} of models with different trade-offs between accuracy and interpretability, instead of a single model.
This gives the user the possibility to inspect multiple options at the end of the run, and make a final call in terms of overall \emph{trustworthiness}, i.e., what model meets \emph{both} acceptable accuracy and interpretability.

Given a supervised learning problem, GP evolves a population of models, here expressed as symbolic functions $f: \mathbb{R}^d \to \mathbb{R}$, such that $f(\vec{x}_i) \simeq y_i, \forall (\vec{x}_i, y_i)$ in the data.
The search of GP is driven by two objectives: accuracy, measured on (part of the) data (another being reserved for validation), and interpretability, measured by the interpretability estimator.
We describe GP further in \Cref{sec:mogp}.
Before that, we describe the neural network used as intepretability estimator (\Cref{sec:ranking-nn}).
This estimator takes as input a vector of six features that describe a model $f$, and produces an output in $\mathbb{R}^2$ that represents the interpretability of $f$ and the respective uncertainty.

Following~\cite{virgolin2020learning}, the six features of $f$ we use for interpretability estimation are: the total size of $f$ (number of variables plus constants plus operations), the number of operations, the number of operations that are non-arithmetic, the maximum length of consecutive compositions of non-arithmetic operations, the number of constants, and the dimensionality of $f$ (i.e., the number of unique variables)---the latter two being an addition with respect to~\cite{virgolin2020learning}.

\subsection{The estimator of interpretability}
\label{sec:ranking-nn}
We use a neural network as interpretability estimator because neural networks have excellent fitting capabilities, can produce uncertainty estimations similarly to Gaussian processes by simply making use of dropout at prediction time~\cite{gal2016dropout}, and their training scales better than that of Gaussian processes~\cite{moore2016fast}.

The neural network takes the 
%(z-score standardized) 
features of a model $f$ as inputs, processes them with three fully-connected hidden layers with \num{100} ReLU activations~\cite{nair2010rectified} and \num{0.25} dropout rate~\cite{srivastava2014dropout}, and finally applies the hyperbolic tangent. 
%(we found this to work well in preliminary testing).
Importantly, we set the network to be trained using binary feedback (i.e., the one provided by the user).
Our training signal is constituted by the features of two models $f_1, f_2$ and a label telling which of the two models is more interpretable for the user.
Given this training signal, we feed the features of $f_1$ and $f_2$ as input to the network, and observe how the label compares to the respective predictions of interpretability $\hat{\psi}_1, \hat{\psi}_2$.
%---for now, for the sake of clarity, let us pretend that these are deterministic, i.e., as if dropout at prediction time is switched off.
Next, we compute a loss that can be seen as a two-points version of the Wasserstein loss commonly used to train generative adversarial networks~\cite{gulrajani2017improved}:
\begin{equation}\label{eq:ranking-net-loss}
    \mathcal{L}(\hat{\psi}_1,\hat{\psi}_2, l) \vcentcolon= l \left( \hat{\psi}_1 - \hat{\psi}_2 \right).
\end{equation}
Here, the value of $l$ depends on the label from the user: $l=-1$ if the first model is deemed more interpretable than the second, else $l=1$.
If, e.g., the network predicts that $\hat{\psi}_1$ is smaller than $\hat{\psi}_2$ but the user prefers $f_1$, then the loss will be positive.
We remark that the proposed loss does not provide an explicit signal on what values $\hat{\psi}$ should assume: this is not an issue because for our model synthesis process, i.e., GP, relative values suffice.
As optimizer we use stochastic gradient descent  ($\text{learning rate}=0.001, \text{momentum}=0.9$).

\paragraph{Active learning with warm-up}
As mentioned  before, the models that are sent to the user for evaluation are those for which the network has registered the maximal prediction uncertainty so far, i.e., while being used to evaluate the models under synthesis.
We measure this uncertainty by applying dropout at prediction time, specifically by taking the standard deviation over repeated predictions $\hat{\psi}^{(1)}, \dots, \hat{\psi}^{(k)}$ for a same input model $f$.
We use $k=10$ as a compromise between speed and fidelity, and set the interpretability estimation $\hat{\psi}$ to be the mean of the $k$ predictions.

To avoid relying solely on the feedback of the user, we include an initial warm-up phase using interpretability estimates generated from the estimator of~\cite{virgolin2020learning} for $100$ random models (from now on, we refer to intepretability estimation according to~\cite{virgolin2020learning} by $\phi$).
We do this because we have no guarantees that the user is responsive, and also so that the initial estimations of interpretability are not completely random.
We heuristically halt the warm-up when the uncertainty of the network starts to decrease, specifically when the standard deviations of the predictions of the last two epochs is smaller than that of the previous two on average. 
Note that an increase of the uncertainty in the initial epochs is a normal consequence of the fact that initialized weights are small (we use Keras default initiliazer~\cite{chollet2015keras}), and as they increase with early gradients, so do the magnitudes of the (still semi-random) predictions. 
We found this scheme to result in around $3$ to $6$ epochs of warm-up.

\paragraph{Justification of adopting a ranking network}
Before proceeding, we provide some evidence that the network we adopt, from now on called a \emph{ranking network}, fares well compared to a more \emph{classic network}, i.e., one trained as a regressor to predict specific values of interpretability.
For the classic network we use the mean squared error loss and use a linear activation to obtain the output (instead of $\tanh$). 
We create a toy scenario where we assume that $\psi := \phi$, i.e., we use the estimator from~\cite{virgolin2020learning} in place of the unknown concept of interpretability of the user, and we create labels for a set of \num{1000} random symbolic models.
Since we use $\phi$ as ground-truth, we change the warm-up phase to use model size ($\ell$) instead of $\phi$ itself, again upon \num{1000} random models.
For the ranking network, we assume that a feedback consists of a label that tells, given \emph{two} symbolic models, which is best (as explained before). 
For the classic network, we have \emph{regression} labels, i.e., the value of $\phi$.
Note that the number of models-per-feedback is twice for the ranking network than for the classic one.
Moreover, providing a label for the classic network requires the user to specify a numeric value, which arguably takes more effort than only telling which of two models is better.
For these reasons, we allow the ranking network to receive double the user's answers than the classic network.

\begin{figure}
    \centering
    %\setlength{\tabcolsep}{0pt}
    %\renewcommand{\arraystretch}{0}
    %\begin{tabular}{cc}
    %    \includegraphics[width=0.5\linewidth]{pics/error_wprioritization.pdf}
    %    &  
    %    \includegraphics[width=0.5\linewidth]{pics/ranking_net_assessment_wpriority_phi_nolabels.pdf}
    %    \\
    %\end{tabular}
    \includegraphics[width=\linewidth]{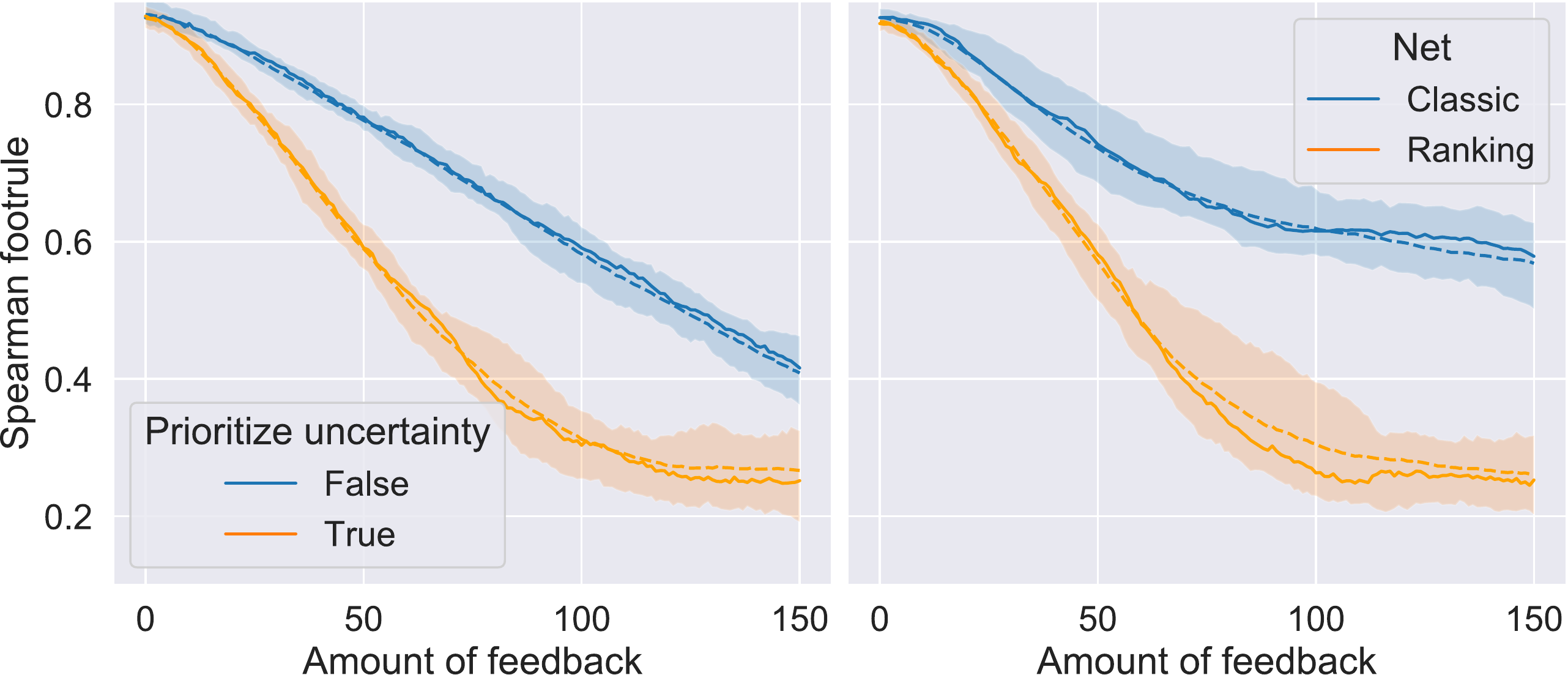}
    \caption{
        Capability of neural networks to learn the same ordering induced by $\phi$ as a function of the amount of feedback (median $=$ solid, mean $=$ dashed, interquartile range $=$ shaded area, across $30$ repetitions).
        A warm-up phase is applied using $\ell$.
        \emph{Left}: Prioritizing uncertainty to decide what models to submit for feedback improves the learning speed of the ranking network. 
        \emph{Right}: The ranking network adapts more quickly than the classic network.
    }
    \label{fig:assessment-ranking-net}
\end{figure}

We consider how well the networks are able to induce the same ranking of $\phi$ with increasing amount of feedback, using the Spearman footrule~\cite{spearman1906footrule,diaconis1977spearman}.
This measures the misalignment between two rankings $R$ and $S$ (in our case, for $r$ models $f_1, \dots, f_r$):
\begin{equation}
    \textit{Spearman footrule}(R,S) = \frac{3}{q(r)} \sum_{i=1}^r |R(f_i) - S(f_i)|,
\end{equation}
where $q(r) = r^2$ is $r$ is even, otherwise $r^2 - 1$.
The term $\frac{3}{q(r)}$ induces a normalization such that the expectation of the footrule is $1$ when the order of one ranking is random with respect to the other, $0$ for identical orders, and $1.5$ for reversed orders.
\Cref{fig:assessment-ranking-net} shows results on this toy setting (with respect to a test set of \num{1000} random models).
On the left panel, it can be seen that the ranking network learns quicker if we ask the user to provide feedback on the models for which the network is most uncertain.
On the right panel, it can be seen that the ranking network outperforms the classic network.
Note that the initial Spearman footrule is lower than $1$ (value for random ranks) thanks to the warm-up phase on $\ell$.

\subsection{Bi-objective evolution}
\label{sec:mogp}
We use the GP version of the well-known NSGA-II algorithm~\cite{deb2000fast} implemented in~\cite{virgolin2020learning}\footnote{Code available at: \url{https://github.com/marcovirgolin/pyNSGP}}.
The models evolved by GP, i.e., the symbolic functions, are represented with a traditional tree-based encoding~\cite{poli2008field}.
At initialization, the trees are generated using the \emph{ramped half-and-half} method, with tree depth between $1$ and $3$.
The trees are rather small, and are allowed to have at most up to $25$ components (nodes) at any point during the evolution, because larger trees quickly becomes incomprehensible.
Trees that are selected with tournaments of size $2$ undergo modification by means of traditional subtree crossover, subtree mutation, and one-point mutation, with equal probability.

Our GP uses two objectives. 
The first objective is maximizing accuracy (for regression, we actually minimize the mean squared error).
We include \emph{linear scaling}~\cite{keijzer2004scaled} to adapt the fitting of the evolving models, since it is often beneficial on real-world datasets~\cite{virgolin2019linear}.
The second objective is maximizing interpretability.
In ML-PIE, this is estimated ($\hat{\psi}$) by our interpretability estimator.
For comparison, we also consider using $\phi$ (estimation according to~\cite{virgolin2020learning}) and $\ell$ (the model size, i.e., number of tree nodes).
The outcome of GP is a collection of models that expresses a \emph{trade-off front} of the objectives, i.e., taken any two models ``from the front'', if the first has better accuracy, then it must have worse interpretability (and vice versa).

The primitives used to evolve symbolic functions are \begin{enumerate*}[label=(\alph*)]
    \item the variables of the problem at hand;
    \item random constants between $-5$ and $5$ with resolution of $0.25$;
    \item the functions $+$, $-$, $\times$, $\div^*$, $\cdot^3$, $\log^*$, $\max$ ($^*$=protected).
\end{enumerate*}
We use a relatively small population of $256$ models and evolve it for $50$ generations.
We found this to be a good compromise between what performance the models can achieve, and how much time the users need to spend to terminate a run of ML-PIE (i.e., approx.~$10$ minutes).
Since the population is not large, we include a simple method to oppose premature convergence that worked remarkably well in preliminary experiments.
Specifically, for each model $f_p$ in the population (parsed in a random order), we check whether another model $f_q$ co-exists that has the same behavior (with respect to the data, i.e., $\forall \vec{x}_i, f_p(\vec{x}_i) = f_q(\vec{x}_i)$); for any such $f_q$, we mark it as duplicate (note, we do not mark the first $f_p$).
Models marked as duplicates are set to have the lowest priority when GP selects parents and chooses the survivors of the generation.

\section{Experimental Setup}
We asked students enrolled at the engineering faculties of the University of Trieste, Italy and Chalmers University of Technology, Sweden to act as users for ML-PIE, which we disseminated as a web application on the cloud.
%We allowed the users to respond for a period of time of approximately two weeks. 
Data collection lasted for approximately two weeks.

Experiments were run on two datasets, namely \emph{Boston housing} (in short, Boston) and \emph{German credit} (German), which concern regression of housing values and binary classification of credit approval, respectively.
These datasets from the UCI repository~\cite{dua2017uci}
are well-known, of moderate-enough size to keep the users busy only for a reasonable time, and also interesting for assessing fairness in AI, because Boston includes a feature regarding race, while German has one regarding sex.
Boston has $506$ examples and $13$ features, while German has \num{1000} examples and $20$ features.
With the parameter settings used for GP, it takes approximately $10$ minutes to complete a run of ML-PIE; the estimation of interpretability by the neural network taking most computation time.
Each run used a random split of the dataset in exam, namely into \SI{70}{\percent} of the examples for training and \SI{30}{\percent} for testing.
The features of the datasets were standardized by z-scoring~\cite{dick2020feature}.

We also prepared an experiment to assess whether ML-PIE is any better than using pre-existing indices of interpretability, namely $\phi$ and $\ell$.
To this end, at the end of each ML-PIE run, the users were presented with a set of models from the trade-off front obtained during the run (i.e., by GP using $\hat{\psi}$) \emph{paired} with models taken from runs where GP used $\phi$ or $\ell$ as indices of interpretability ($100$ pre-computed runs for each).
Models in each pair ($\hat{\psi}$ vs.~$\phi$ or $\hat{\psi}$ vs.~$\ell$) had similar accuracy (pairing made by considering the minimal absolute difference).
The users were asked to tell, for each pair, which of the two models was more interpretable.
The users were not told which models were found during their session (i.e., with $\hat{\psi}$) and which were found using $\phi$ or $\ell$, to prevent bias.

\section{Results}
We show key results regarding the workings of \name{} and the performance it achieves.
A total of $61$ users answered the call, leading to $49$ runs for Boston and $45$ runs on German (some students did not run both).
In the following, we report: how the users and the estimator of interpretability behaved during the runs (\Cref{sec:user-estim-behavior}); the search performance of ML-PIE compared to that obtainable when $\phi$ or $\ell$ are used (\Cref{sec:search-performance}); whether estimations of interpretability $\hat{\psi}$ obtained by different users behave differently (\Cref{sec:assessment-personalization}); and, last but not least, whether the users preferred models discovered using $\hat{\psi}$ over models found by $\phi$ or $\ell$ (\Cref{sec:final-survey}).

\subsection{Behavior of the users and of the interpretability estimator}\label{sec:user-estim-behavior}
\begin{figure}
    \centering
    \setlength{\tabcolsep}{0pt}
    \renewcommand{\arraystretch}{0}
    \begin{tabular}{l}
        \includegraphics[width=\linewidth]{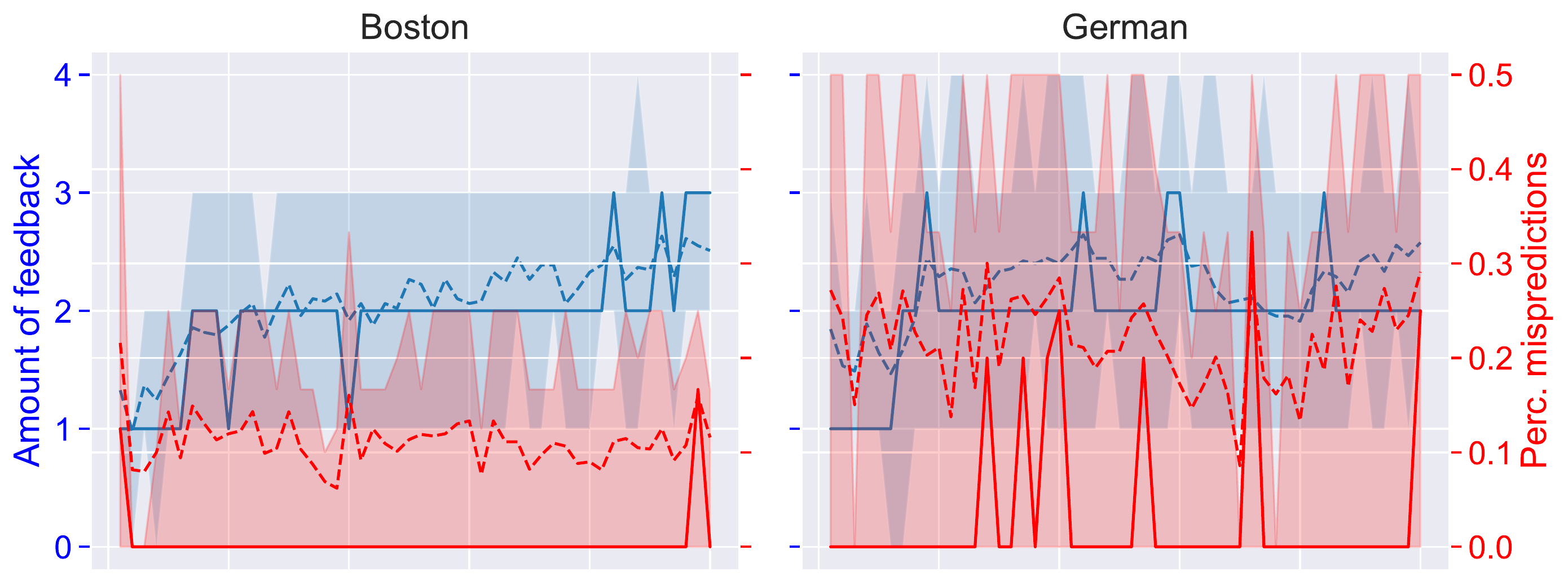} \\ \includegraphics[width=\linewidth]{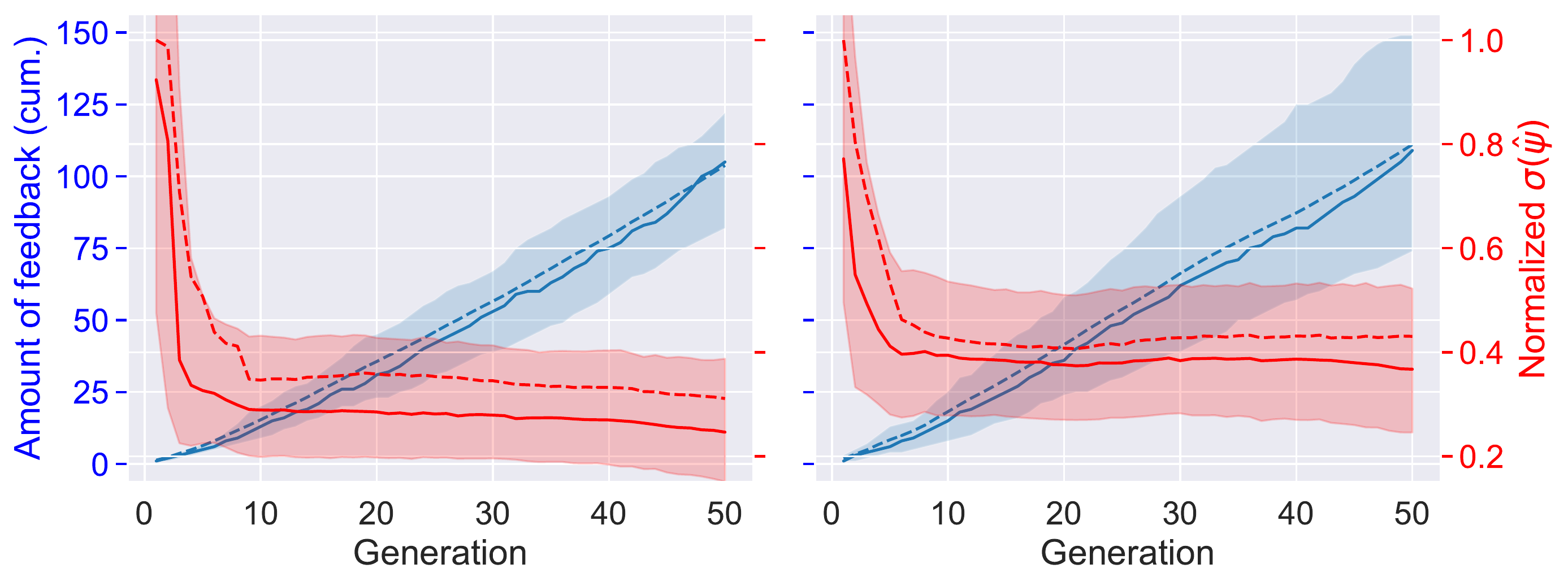}
    \end{tabular}
    \caption{
        User and framework behavior over the number of generations (median $=$ solid, mean $=$ dashed, interquartile range $=$ shaded area, across all users). 
        \emph{Top}: Amount of feedbacks given and percentage of mispredictions made by the intepretability estimator, per generation.
        \emph{Bottom}: Cumulative amount of feedback and estimator uncertainty $\sigma(\hat{\psi})$ over the generations. 
        Uncertainty is normalized by dividing by the average uncertainty observed at the first generation.
    }
    \label{fig:feedbacks-stats-per-gen}
\end{figure}

\Cref{fig:feedbacks-stats-per-gen} summarizes how the users and the estimator behaved over time.
In the top panel, we see that the amount of feedback given per generation tends to be slightly smaller for early generations. 
We remark that this is because, in GP, the average model size tends to grow over time, causing early generations to be computed more quickly than later ones, and the users to have less time to provide feedback.
We also measured the rate at which the estimator made mispredictions, by checking whether the estimates $\hat{\psi}_1, \hat{\psi}_2$ for the respective models $f_1$ and $f_2$ \emph{before the user's answer} turned out to be in disagreement with the user's answer (e.g., $\hat{\psi}_1 < \hat{\psi}_2$ but the user picks $f_1$).
Mispredictions were quite rare. 
This result is in line with what can be seen on the bottom panel, i.e., that the uncertainty of the estimator (the standard deviation of its estimations $\sigma(\hat{\psi})$) dropped rather quickly and remained low over time.
We also see that the cumulative amount of feedback had a steady, linear growth. 

\Cref{tab:example-models} shows examples of models\footnote{We do not perform model simplification because too costly to perform upon each model being evolved, it might harm evolvability, and, if done only for the models displayed to the user, it may create a perception mismatch with respect to the features of the models (size, no.~operators, etc.), used by the interpretability estimator.} extracted from the trade-off fronts obtained by the user-guided evolutions, at different trade-off levels between accuracy and interpretability (interpretability percentile $\tau$). 
Note that we actually report errors: the training and test errors for Boston are mean squared errors, while those for German are inaccuracies ($1-\text{accuracy}$).
It can clearly be seen that, the more we move from more accurate models to more interpretable ones (i.e., increasing $\tau$), the more the models become simpler under several aspects, such as total size, number of dimensions, and presence of non-arithmetic operations.
This confirms that the estimator was capable of capturing a sensible notion of interpretability.
A further confirmation of this is provided later in \Cref{sec:assessment-personalization}.

\textbf{Take-home.} Overall, the users provided good amounts of feedback through the process, and the estimator of interpretetability responded in a sensible manner.
Observed uncertanties, mispredictions, and model ranking behaviors, are reasonable.

\begin{table*}
    \centering
    \caption{
        Examples of models discovered by evolutions guided by the users.
        The models are taken at random from the evolved trade-off fronts (training error vs.~$\hat{\psi}$) at different interpretability percentiles ($\tau$).
        The test error is also reported, and some amount of overfitting can be observed (most notably, the first model at $\tau=10$ for Boston).
        Our approach is capable of capturing sensible trends of model intepretability: models become simpler as $\tau$ increases.
    }
    \begin{tabular}{
    c
    c
    S[table-format=1.2]
    S[table-format=1.2]
    l}
\toprule
    Dataset & $\tau$ & {Train} & {Test} & {Model} \\
\midrule
\multirow{9}{*}{\rotatebox[origin=c]{90}{Boston}} & \multirow{3}{*}{10} & 18.43 & 37.53 & $x_{12}-\operatorname{max}\left(2.5, x_{8}+x_{9}+x_{5}-x_{8}-x_{2}\right) \left(-0.25+x_{12}\right)+\operatorname{max}\left(x_{5}, x_{12}\right)$ \\
       %& & 22.23 & 14.13 & $-0.75\times\left(x_{5} \left(x_{9}-3.5+x_{9}\right)-3.5\right) x_{12}+x_{12}+x_{12}$ \\
       & & 19.86 & 18.57 & $4.5\times x_{12}+x_{5} \left(x_{10}+\frac{x_{9}}{x_{5}}\right)+\left(x_{12}+x_{10}\right) x_{5}+x_{12}+x_{5} \left(-3.25-x_{5}\right)$ \\
       %& & 24.17 & 17.47 & $x_{10}+x_{5}+x_{5} x_{10}+x_{10}+x_{12}-\operatorname{max}\left(x_{6}, x_{5}+x_{5}\right)+x_{12}-x_{5}+x_{12}$ \\
       & & 21.24 & 20.07 & $\operatorname{max}\left(x_{11} x_{11} x_{11}, \ln(\left|x_{5} \left(x_{5}-x_{12}\right)\right|)\right)-x_{12}+x_{5}-\operatorname{max}\left(x_{0}, 2.5\right)+x_{12}+x_{10}$ \\
       \cmidrule(lr){3-5}
%       & \multirow{5}{*}{30} & 23.50 & 22.03 & $x_{12}+x_{12}-x_{5}+x_{0}-x_{5} x_{5}$ \\
%       & & 22.56 & 40.84 & $x_{12}-x_{5}+x_{5} x_{2}$ \\
%       & & 24.40 & 20.87 & $\operatorname{max}\left(x_{5}, x_{0}-x_{0}\right) x_{5}-x_{12}+x_{0}+x_{12}+x_{12}+x_{10}-x_{5}+x_{12}$ \\
%       & & 21.35 & 22.41 & $4.5\times x_{12}+x_{5}+x_{5}+x_{10}+x_{5} \left(-3.25-x_{5}\right)$ \\
%       & & 20.24 & 23.42 & $\operatorname{max}\left(x_{5}-x_{12}, \frac{x_{7}}{x_{7}}\right)-x_{12}$ \\
%       \cmidrule(lr){3-5}
       & \multirow{3}{*}{50} & 25.64 & 30.16 & $x_{12}-\operatorname{max}\left(x_{12}, x_{5}\right)$ \\
       & & 26.05 & 35.42 & $\frac{-1.0+x_{4}}{4.25}-4.75-1.75-\left(x_{12}+x_{10}-x_{5}\right)\times4.75$ \\
       & & 32.13 & 29.09 & $\frac{x_{12}}{x_{12}} \left(x_{12}+x_{12}-x_{5}\right)$ \\
       % & & 16.25 & 37.56 & $x_{2}+x_{12}-\left(x_{5}+2.75\right) x_{5}+x_{10}+x_{12}$ \\
       %& & 31.33 & 29.09 & $\frac{x_{5}}{1.5}-x_{12}$ \\
       \cmidrule(lr){3-5}
%       & \multirow{5}{*}{70} & 29.09 & 42.56 & $x_{5}+x_{5}-x_{12}$ \\
%       & & 30.57 & 53.34 & $x_{9}-x_{5}+x_{5}$ \\
%       & & 28.26 & 26.23 & $x_{12}+x_{12}+x_{10}-x_{5}$ \\
%       & & 32.00 & 37.01 & $x_{12}-\frac{x_{5}}{4.5}$ \\
%       & & 28.37 & 25.80 & $x_{4}-x_{5}-3.25-x_{12}+\frac{x_{5}}{-2.0}-x_{12}-x_{12} x_{5}+x_{12}$ \\
%       \cmidrule(lr){3-5}
       & \multirow{3}{*}{90} & 35.47 & 45.54 & $\frac{x_{12}}{-2.75}\times-0.25$ \\
       & & 57.82 & 61.84 & $x_{10}-x_{11}$ \\
       %& & 46.94 & 40.11 & $\frac{x_{12}}{x_{3}}$ \\
       & & 71.97 & 62.50 & $3.0-x_{4}$ \\
       %& & 82.43 & 73.38 & $x_{5} x_{9}$ \\
\cline{1-5}
\multirow{9}{*}{\rotatebox[origin=c]{90}{German}} & \multirow{3}{*}{10} & 0.24 & 0.26 & $x_{1}-x_{0}-x_{6}+x_{7}+x_{11}-x_{6}+\frac{x_{1}}{x_{17}}-x_{2}-x_{0}$ \\
       %& & 0.23 & 0.25 & $x_{7}-\operatorname{max}\left(x_{2}, x_{0}\right)-x_{0}-x_{1}-x_{5}-\operatorname{max}\left(x_{4}, x_{5}\right)-x_{0}$ \\
       %& & 0.24 & 0.26 & $x_{1}-x_{0}+x_{2}+\frac{x_{16}-x_{6}}{\frac{-2.0}{x_{6}}}+x_{0}$ \\
       & & 0.25 & 0.24 & $x_{4}+x_{19}-x_{10}+\frac{x_{0}}{x_{16}}-x_{1}+x_{18}-x_{1}-x_{1}+x_{5}-x_{6}-x_{6}$ \\
       & & 0.25 & 0.25 & $x_{2}+x_{2}+x_{8}+x_{0}-x_{1}$ \\
       \cmidrule(lr){3-5}
%       & \multirow{5}{*}{30} & 0.23 & 0.28 & $x_{1}-x_{2}-x_{5}-x_{0}-x_{7}$ \\
%       & & 0.26 & 0.27 & $\frac{x_{0}+x_{2}}{x_{4}+4.75}$ \\
%       & & 0.25 & 0.26 & $x_{0}+\frac{x_{4}-x_{2}}{-3.5}$ \\
%       & & 0.24 & 0.31 & $x_{18}+x_{0}-x_{11}-x_{2}+x_{0}$ \\
%       & & 0.25 & 0.25 & $\left(x_{0}+x_{0}-x_{1}\right) \left(3.0-x_{5}\right)$ \\
%       \cmidrule(lr){3-5}
       & \multirow{3}{*}{50} & 0.25 & 0.24 & $\left(x_{2}-x_{4} x_{4}+x_{1}-x_{4}-4.5\right) x_{0}$ \\
       %& & 0.27 & 0.24 & $x_{11}-x_{0}+x_{2}$ \\
       %& & 0.24 & 0.23 & $x_{2}-x_{7}-x_{0}-x_{1}-x_{0}$ \\
       & & 0.25 & 0.27 & $x_{11}-2.5-x_{2}-x_{0}+3.25$ \\
       & & 0.26 & 0.30 & $x_{4}-x_{0}-x_{0}$ \\
       \cmidrule(lr){3-5}
%       & \multirow{5}{*}{70} & 0.27 & 0.25 & $x_{0}+x_{0}-x_{1}$ \\
%       & & 0.28 & 0.29 & $\frac{x_{0}}{x_{16}}$ \\
%       & & 0.28 & 0.29 & $\frac{x_{0}}{x_{16}}$ \\
%       & & 0.27 & 0.29 & $x_{0}+x_{2}$ \\
%       & & 0.26 & 0.30 & $x_{5}-x_{1}$ \\
%       \cmidrule(lr){3-5}
       & \multirow{3}{*}{90} & 0.26 & 0.29 & $x_{11}-x_{0}$ \\
       & & 0.26 & 0.30 & $x_{0}-x_{4}+x_{5}$ \\
       %& & 0.30 & 0.31 & $\frac{x_{6}}{x_{1}}$ \\
       %& & 0.26 & 0.29 & $x_{0}-x_{11}$ \\
       & & 0.28 & 0.26 & $x_{2}+x_{0}$ \\
\bottomrule
\end{tabular}
    \label{tab:example-models}
\end{table*}

\subsection{Search performance}\label{sec:search-performance}
\Cref{tab:search-performance} summarizes the search performance of GP using $\hat{\psi}$ (over the runs performed during the feedback sessions with the users, $49$ for Boston and $45$ for German), $\phi$, and $\ell$ ($100$ runs per dataset for each).
Before delving into $\hat{\psi}$, we remark that the results obtained by $\phi$ and $\ell$ are similar to those reported in literature (see, e.g.,~\cite{virgolin2020learning} for Boston\footnote{Besides front sizes, this can be checked by de-normalizing the errors reported in Table~2 of~\cite{virgolin2020learning} by multiplying them with the variance of the label, i.e., $\approx 84.6$.} and~\cite{mota2021towards,setzu2021glocalx} for German), despite the fact that our GP used limited resources ($\text{population size}=256,\text{generations}=50$). 
%Albeit other differences exist between this work and~\cite{virgolin2020learning} (cfr.~function sets and rates of variation), we identified the penalization of duplicate models to avoid premature convergence to be the main cause of improvement.
%Search using $\ell$ benefits the most of this. 

What can be mainly drawn from \Cref{tab:search-performance} is that, error-wise, results with $\hat{\psi}$ can be slightly worse than those obtained with $\phi$ and $\ell$.
However, this is rarely important (e.g., only statistically significant in two cases at test time, for $\tau=50$). 
Likely, this is a consequence of the fact that the estimator of $\hat{\psi}$ is stochastic (due to dropout) and is only refined over time. 
Noisy objectives are known to hinder search algorithms, including evolutionary ones~\cite{beyer2000evolutionary,rakshit2017noisy}.
This fact also justifies larger front sizes being produced by $\hat{\psi}$. 
We note that if we restrict the results with $\hat{\psi}$ to only account for good amounts of feedback being given, e.g., $\geq 100$ (approximately \SI{50}{\percent}), results do not change substantially.

\textbf{Take-home.} 
Model accuracy is slightly worse when $\hat{\psi}$ is used instead of $\phi$ or $\ell$.
This is likely because the estimator is noisy. 
A simple action that can be taken is to increase GP search resources.

\begin{table}[]
    \centering
    \begin{tabular}{
    r @{\hskip 3mm} r
    S[table-format=2.1] @{\hskip 3mm} S[table-format=2.1] @{\hskip 3mm} S[table-format=2.1] @{\hskip 3mm} S[table-format=2.1]
    S[table-format=1.2] @{\hskip 3mm} S[table-format=1.2] @{\hskip 3mm} S[table-format=1.2] @{\hskip 3mm} S[table-format=1.2]
    }
    \toprule
    \multicolumn{2}{r}{Dataset} & \multicolumn{4}{c}{Boston} & \multicolumn{4}{c}{German}\\
     \cmidrule(lr){3-6}
     \cmidrule(lr){7-10}
    \multicolumn{2}{r}{$\tau$} & {0} & {10} & {30} & {50} & {0} & {10} & {30} & {50} \\
    \midrule
    \multirow{3}{*}{$\hat{\psi}$}
    & Front & \multicolumn{4}{c}{ 12.4 } & \multicolumn{4}{c}{ 10.9 } \\
    & Train & 20.06 & 21.06 & 23.32 & 27.78 & 0.23 & 0.24 & 0.25 & 0.26 \\
    & Test & 22.43 & 23.19 & 24.65 & 28.93 & 0.26 & 0.26 & 0.27 & 0.27 \\
    \midrule
    \multirow{3}{*}{$\phi$}
    & Front & \multicolumn{4}{c}{ 9.3 } & \multicolumn{4}{c}{ 5.4 } \\
    & Train & 19.09 & \bfseries 19.39 & \bfseries 20.94 & \bfseries 23.24 & 0.24 & 0.24 & 0.24 & 0.25 \\
    & Test & 22.62 & 21.92 & 22.44 & \bfseries 24.98 & 0.27 & 0.27 & 0.27 & 0.28 \\
    \midrule
    \multirow{3}{*}{$\ell$}
    & Front & \multicolumn{4}{c}{ 9.1 } & \multicolumn{4}{c}{ 6.9 } \\
    & Train & 18.70 & \bfseries 19.03 & \bfseries 20.54 & \bfseries 22.35 & 0.24 & 0.24 & 0.24 & 0.25 \\
    & Test & 22.39 & 22.68 & 23.08 & \bfseries 24.24 & 0.27 & 0.27 & 0.27 & 0.28 \\
    \bottomrule
    \end{tabular}
    \caption{
        Mean front sizes (Front), and mean train and test errors (respectively Train and Test), at different interpretability percentiles ($\tau$) of the models in the best-found trade-off fronts, for GP using $\hat{\psi}$, $\phi$, or $\ell$ as interpretablity indices.
        Errors in bold are significantly better than their $\hat{\psi}$ counterpart (Mann-Whitney U test $p\text{-value} < 0.01$ with Bonferroni correction).
    }
    \label{tab:search-performance}
\end{table}

\subsection{Level of personalization}\label{sec:assessment-personalization}
\begin{figure}
    \centering
    \includegraphics[width=\linewidth]{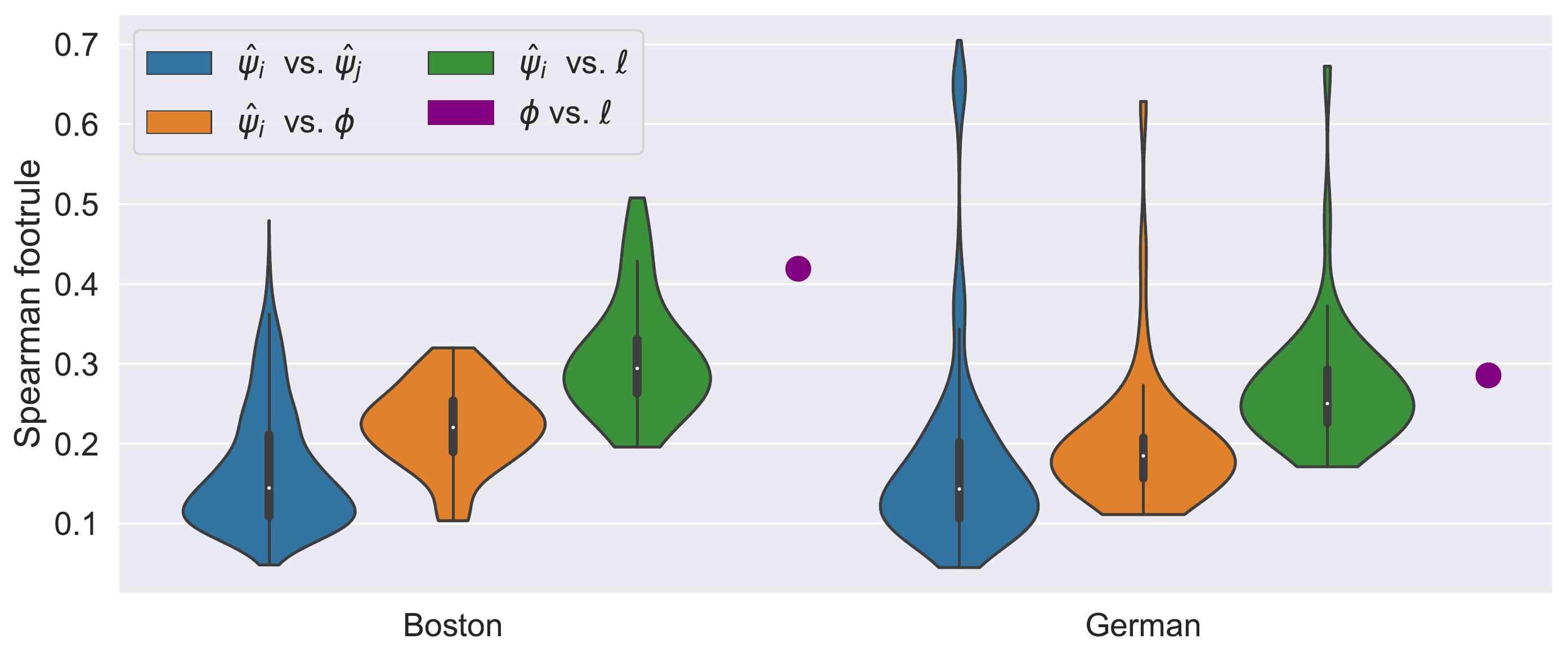}
    \caption{
        Violin plot of Spearman footrules from comparing rankings obtained with different interpretability estimators.
        On average, a user's estimated interpretability $\hat{\psi}_i$ is more similar to that of another user $\hat{\psi}_j$ than to $\phi$ and $\ell$.
        However, high disagreements are also possible (fat and long tail for Boston, very long tail for German).
    }
    \label{fig:trained-network-distributions}
\end{figure}

\Cref{fig:trained-network-distributions} shows comparisons of rankings induced by different users with respect to each other and with respect to $\phi$ and $\ell$.
We now use the notation $\hat{\psi}_i$ to refer to the interpretability estimates for a certain $i$-th user.
The rankings are computed by ordering, according to $\phi$, $\ell$, and each $\hat{\psi}_i$, all models from the fronts ever discovered by all the users (excluding syntactic duplicates and totalling $\approx \num{9000}$ models).  

We begin by observing that the distributions of Spearman footrule obtained when comparing user induced-rankings with one another ($\hat{\psi}_i$ vs.~$\hat{\psi}_j$) can have heavy or long tails.
Overall, this corroborates the hypothesis that some users may have non-negligible disagreements regarding their notions of interpretability, and thus that seeking personalized interpretability estimations is needed.

Average behaviors are also interesting.
On average, user induced-rankings are more similar to one another (footrule of $\hat{\psi}_i$ vs.~$\hat{\psi}_j$ has average value of $\approx 0.13$) than to those induced by $\phi$ or $\ell$ (respectively $\hat{\psi}_i$ vs.~$\phi$ with average value $\approx 0.2$ and $\hat{\psi}_i$ vs.~$\ell$ with average value $\approx 0.25$).
This means that, on average, users provided similar feedback, i.e., they agreed on a similar notion of interpretability $\psi$ (at least, for the experimental setting we proposed and for the users involved).
%Another fact to consider is that there may exist biases in the training, i.e., the networks may tend, on average, to learn something similar even if feedbacks are somewhat different.
%We cannot exclude this factor but we believe that, probably, both facts (many user's agree on $\psi$ and there is some inherent learning bias) are at play, to some extent.
Furthermore, note that user induced-rankings are more similar to those of $\phi$ and $\ell$ than how the ranking of $\phi$ is similar to that of $\ell$. 
This means that the average $\hat{\psi}_i$ has more in common with $\phi$ and $\ell$ than the latter two have with each other.

\textbf{Take-home.} The estimations can be personalized (long tails of $\hat{\psi}_i$ vs.~$\hat{\psi}_j$) but are similar on average (footrule of $\approx 0.13$), conditioned to our experimental settings and involved cohort (students).

\subsection{Final user judgment on $\hat{\psi}$, $\phi$, and $\ell$}\label{sec:final-survey}
\begin{figure}
    \centering
    \setlength{\tabcolsep}{0pt}
    \renewcommand{\arraystretch}{0}
    \begin{tabular}{c}
        \includegraphics[width=\linewidth]{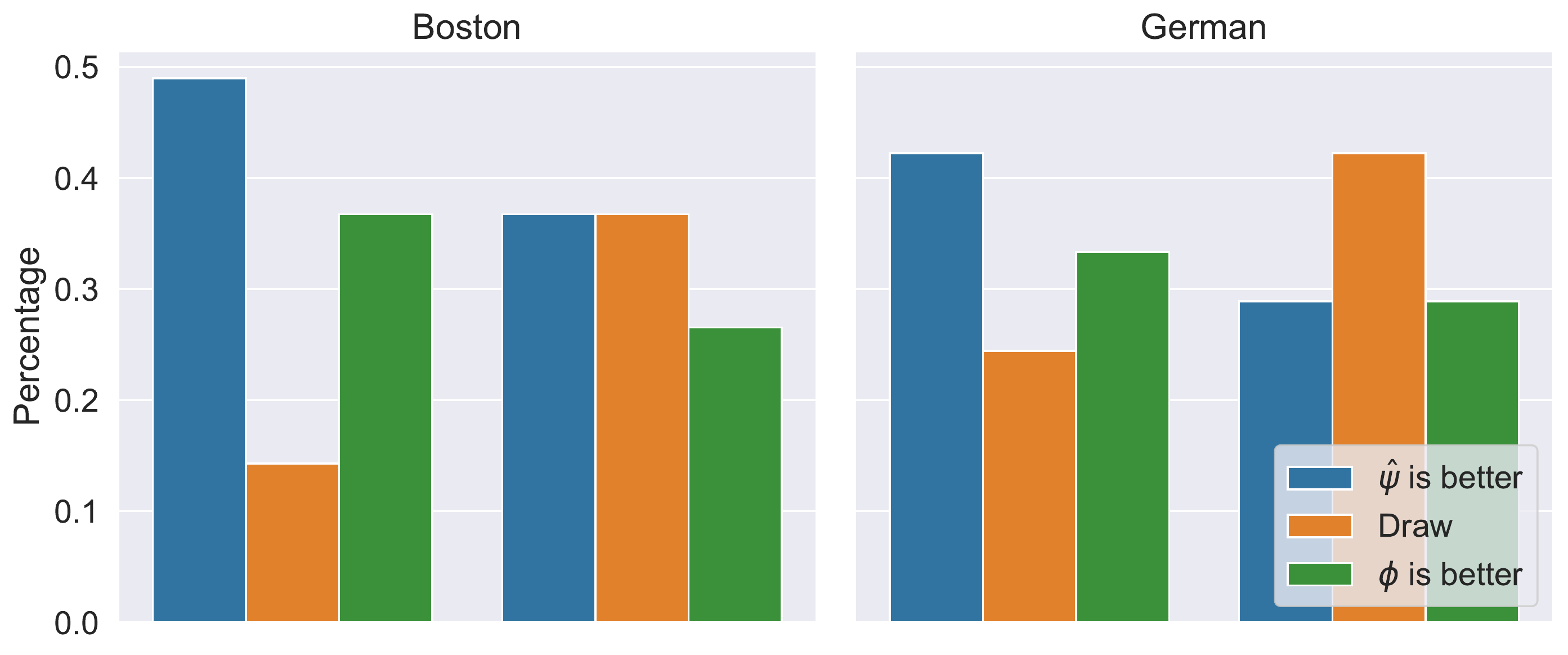}
        \\
        \includegraphics[width=\linewidth]{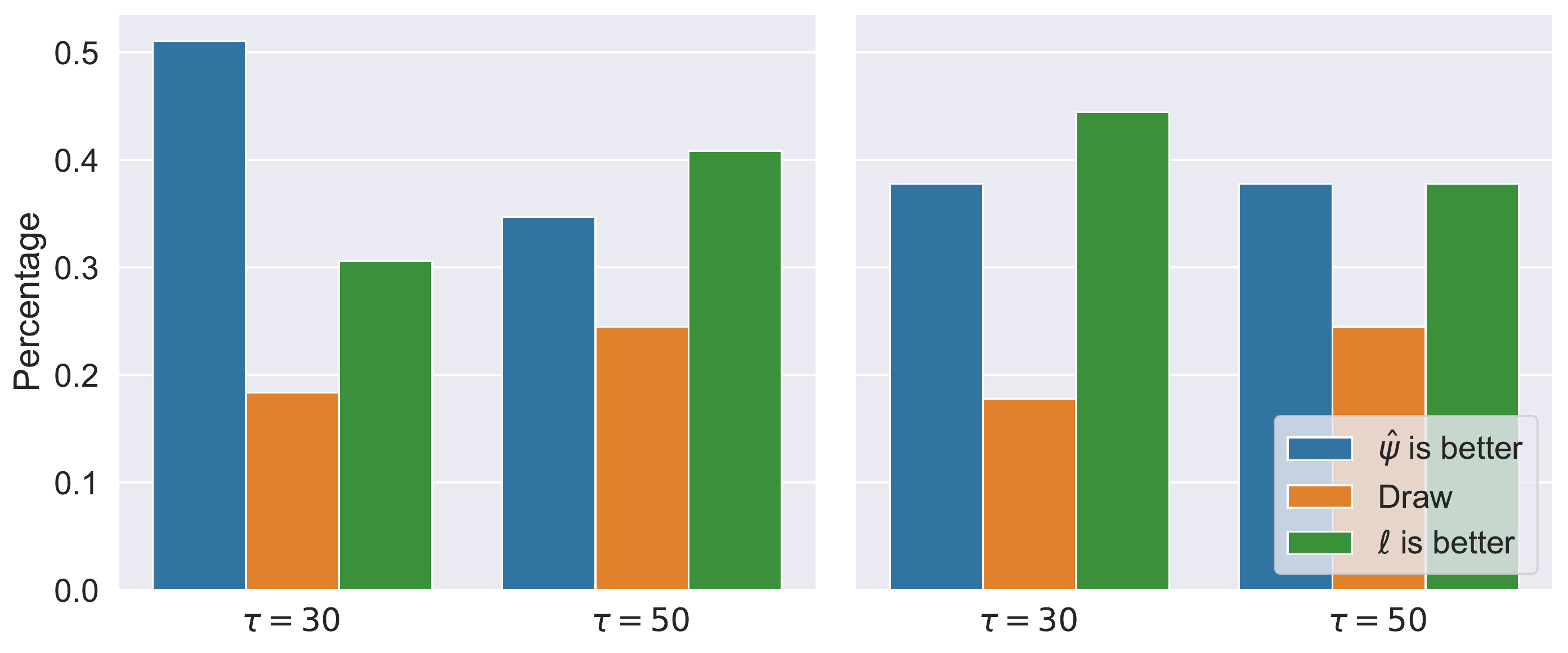}
    \end{tabular}
    \caption{
        Outcome of the survey at the end of the evolution.
        For $\tau=50$, it largely does not matter what index of interpretability is used. 
        When the models start to be more complex, i.e., for $\tau=30$, the users generally prefer $\hat{\psi}$ over $\phi$ and $\ell$, except for $\hat{\psi}$ vs.\ $\ell$ on German.
    }
    \label{fig:final-survey-results}
\end{figure}

\Cref{fig:final-survey-results} shows the preferences indicated by the users when presented, at the end of an ML-PIE run, with models found by the run itself at $\tau=30$ and $\tau=50$, and similarly accurate models obtained in pre-performed runs using $\phi$ and $\ell$.

The first result we highlight is that, for sufficiently simple models (i.e., $\tau=50$, cfr.~\Cref{tab:example-models}), it does not seem to matter what interpretability index is used.
Conversely, preferences start to emerge when more complex models are at play, i.e., at $\tau=30$.
In $\hat{\psi}$ vs.~$\phi$ comparisons (top panels), we see that models discovered by our approach were preferred over those found using $\phi$ almost \SI{15}{\percent} of the times for Boston, and \SI{10}{\percent} of the times for German.
The biggest difference at $\tau=30$ happens for $\hat{\psi}$ vs.~$\ell$ for Boston (bottom left panel), with a \SI{20}{\percent} preference gain for $\hat{\psi}$.
For German at $\tau=30$, however, $\ell$ was preferred over $\hat{\psi}$ by approximately \SI{7}{\percent}.
We can justify this result by looking back at \Cref{tab:example-models}, where it can be sees that the more complex models found for German ($\tau=10$) are actually not that complex after all (e.g., they all lack non-arithmetic operations).
On this dataset, decent accuracy can already be obtained when the models are linear, hence minimizing size seems to be sufficient.
Note that, by contrast, the models found for Boston can be much more complex.

\textbf{Take-home.} Overall, for more complex models, the users indicated a preference for the models found with ML-PIE over those found using $\phi$ or $\ell$.
However, for simpler models, $\ell$ suffices.

\section{Discussion \& conclusion}\label{sec:discussion}
We believe that our results, obtained on two datasets and involving $61$ engineering students, are promising.
The results show that ML-PIE behaves sensibly, the search performance of GP is not harmed excessively by $\hat{\psi}$ being noisy, different users can induce different $\hat{\psi}$, and the models obtained by ML-PIE tend to be preferred over those obtained when $\phi$ or $\ell$ are used instead.

Regarding the last point, preference for $\hat{\psi}$ over $\phi$ and $\ell$ is, however, only marginal. 
We believe that this result is a consequence of what can be considered the main limitation of this study: we did not involve actual domain experts and used a rather general set of model features to estimate model interpretability.
%Our users, i.e., engineering students, could only base their choices upon factors such as the concatenation of non-arithmetic functions, the presence of multiple dimensions, and so on.
Since our users (engineering students) were not experts in housing values (for Boston) nor credit approval risk (for German), we did not provide them with the meaning of the features of the datasets, nor we asked them to attempt to understand what the models in exam exactly mean.
Yet, relying on students allowed us to have a decently-sized cohort with some general knowledge about formulae: we believe that this allowed us to gain sufficient evidence that ML-PIE works sensibly.
We think that even if the preference gains obtained by $\hat{\psi}$ over $\phi$ and $\ell$ were moderate, they represent a very promising result: 
by enabling the interpretability estimator to use more specialized model features (e.g., what domain-specific features and functions appear, how many times, and in which context), and involving actual domain experts,
%to assess the models in detail, 
it is reasonable to expect that the preference for $\hat{\psi}$ will increase substantially.

From a technical standpoint, there are a number of aspects worth studying to improve ML-PIE.
For example, while we used a neural network with sensible settings to realize the interpretability estimator, we did not, e.g., optimize its architecture.
Another point that deserves further investigation is concurrency aspects. 
We designed ML-PIE to make the user provide feedback while the model synthesis process is ongoing because, this way, the models that require user assessment come directly from the distribution of models under synthesis---and not from a potentially-unrepresentative sample, wasting (part of) the user's effort.
However, the pace of feedback provided by the user may be too slow with respect to the speed of the model synthesis process, ultimately harming it.
It may be beneficial to study whether the model synthesis process should be halted at times, e.g., based on statistics regarding the uncertainty of the interpretability estimator.

In the future, together with involving domain experts and adding representative power to the estimator of interpretability (e.g., enabling more model features to be accounted for), we plan to demonstrate that ML-PIE can be adapted to other settings than formulae-like models (e.g., decision trees) and other model synthesis processes than evolutionary ones (e.g., single-objective reinforcement learning or multi-objective gradient descent~\cite{deist2021multi}).
Regarding the last point, it would be interesting to study whether model synthesis can be guided exclusively by user feedback, to optimize some overall measure of \emph{trust}~\cite{ferrario2019ai}, e.g., a measure that encapsulates, at once, accuracy, interpretability, and possibly other aspects such as soundness (e.g., for models that must satisfy physical constraints).

Concluding, we propose Model Learning with Personalized Interpretability Estimation (ML-PIE), a human-in-the-loop, active learning approach for the synthesis of models with user-tailored interpretability.
ML-PIE answers the need for eXplainable AI (XAI) methods capable of synthesizing models that have a chance of being interpretable at a personalized level.
Our experiments involving $61$ students of engineering show that ML-PIE is capable of learning, from their feedback, user-specific estimations of interpretability.
These estimations can be substantially different for different users, and are in general different from interpretability indices from literature.
Last but not least, the users show a moderate preference for the models synthesized using their own feedback over models synthesized when other interpretability indices are used.

We believe that this work shows that XAI methods capable of model synthesis with personalized interpretability can be achieved.
These methods may bring profound advantages for the responsible use of AI in high-stakes societal applications.

%\begin{acks}
%Do we need to thank anybody?
%\end{acks}

\bibliographystyle{ACM-Reference-Format}
\bibliography{references}


\begin{thebibliography}{63}

%%% the Web address.

\ifx \showCODEN    \undefined \def \showCODEN     #1{\unskip}     \fi
\ifx \showDOI      \undefined \def \showDOI       #1{#1}\fi
\ifx \showISBNx    \undefined \def \showISBNx     #1{\unskip}     \fi
\ifx \showISBNxiii \undefined \def \showISBNxiii  #1{\unskip}     \fi
\ifx \showISSN     \undefined \def \showISSN      #1{\unskip}     \fi
\ifx \showLCCN     \undefined \def \showLCCN      #1{\unskip}     \fi
\ifx \shownote     \undefined \def \shownote      #1{#1}          \fi
\ifx \showarticletitle \undefined \def \showarticletitle #1{#1}   \fi
\ifx \showURL      \undefined \def \showURL       {\relax}        \fi
% invisible to TeX
\providecommand\bibfield[2]{#2}
\providecommand\bibinfo[2]{#2}
\providecommand\natexlab[1]{#1}
\providecommand\showeprint[2][]{arXiv:#2}

\bibitem[\protect\citeauthoryear{Adadi and Berrada}{Adadi and Berrada}{2018}]%
        {adadi2018peeking}
\bibfield{author}{\bibinfo{person}{Amina Adadi} {and} \bibinfo{person}{Mohammed
  Berrada}.} \bibinfo{year}{2018}\natexlab{}.
\newblock \showarticletitle{Peeking inside the black-box: {A} survey on
  e{X}plainable {A}rtificial {I}ntelligence ({XAI})}.
\newblock \bibinfo{journal}{\emph{IEEE Access}}  \bibinfo{volume}{6}
  (\bibinfo{year}{2018}), \bibinfo{pages}{52138--52160}.
\newblock


\bibitem[\protect\citeauthoryear{Bartoli, De~Lorenzo, Medvet, and
  Tarlao}{Bartoli et~al\mbox{.}}{2017}]%
        {bartoli2017active}
\bibfield{author}{\bibinfo{person}{Alberto Bartoli}, \bibinfo{person}{Andrea
  De~Lorenzo}, \bibinfo{person}{Eric Medvet}, {and} \bibinfo{person}{Fabiano
  Tarlao}.} \bibinfo{year}{2017}\natexlab{}.
\newblock \showarticletitle{Active learning of regular expressions for entity
  extraction}.
\newblock \bibinfo{journal}{\emph{IEEE Transactions on Cybernetics}}
  \bibinfo{volume}{48}, \bibinfo{number}{3} (\bibinfo{year}{2017}),
  \bibinfo{pages}{1067--1080}.
\newblock


\bibitem[\protect\citeauthoryear{Benk and Ferrario}{Benk and Ferrario}{2020}]%
        {benk2020explaining}
\bibfield{author}{\bibinfo{person}{Michaela Benk} {and} \bibinfo{person}{Andrea
  Ferrario}.} \bibinfo{year}{2020}\natexlab{}.
\newblock \bibinfo{title}{Explaining Interpretable Machine Learning: {T}heory,
  Methods and Applications}.
\newblock
\newblock
\newblock
\shownote{SSRN Methods and Applications.}


\bibitem[\protect\citeauthoryear{Beyer}{Beyer}{2000}]%
        {beyer2000evolutionary}
\bibfield{author}{\bibinfo{person}{Hans-Georg Beyer}.}
  \bibinfo{year}{2000}\natexlab{}.
\newblock \showarticletitle{Evolutionary algorithms in noisy environments:
  {T}heoretical issues and guidelines for practice}.
\newblock \bibinfo{journal}{\emph{Computer Methods in Applied Mechanics and
  Engineering}} \bibinfo{volume}{186}, \bibinfo{number}{2-4}
  (\bibinfo{year}{2000}), \bibinfo{pages}{239--267}.
\newblock


\bibitem[\protect\citeauthoryear{Breslow and Aha}{Breslow and Aha}{1997}]%
        {breslow1997simplifying}
\bibfield{author}{\bibinfo{person}{Leonard~A. Breslow} {and}
  \bibinfo{person}{David~W. Aha}.} \bibinfo{year}{1997}\natexlab{}.
\newblock \showarticletitle{Simplifying decision trees: {A} survey}.
\newblock \bibinfo{journal}{\emph{Knowledge engineering review}}
  \bibinfo{volume}{12}, \bibinfo{number}{1} (\bibinfo{year}{1997}),
  \bibinfo{pages}{1--40}.
\newblock


\bibitem[\protect\citeauthoryear{Cano, Zafra, and Ventura}{Cano
  et~al\mbox{.}}{2013}]%
        {cano2013interpretable}
\bibfield{author}{\bibinfo{person}{Alberto Cano}, \bibinfo{person}{Amelia
  Zafra}, {and} \bibinfo{person}{Sebasti{\'a}n Ventura}.}
  \bibinfo{year}{2013}\natexlab{}.
\newblock \showarticletitle{An interpretable classification rule mining
  algorithm}.
\newblock \bibinfo{journal}{\emph{Information Sciences}}  \bibinfo{volume}{240}
  (\bibinfo{year}{2013}), \bibinfo{pages}{1--20}.
\newblock


\bibitem[\protect\citeauthoryear{Chollet et~al\mbox{.}}{Chollet
  et~al\mbox{.}}{2015}]%
        {chollet2015keras}
\bibfield{author}{\bibinfo{person}{Fran\c{c}ois Chollet} {et~al\mbox{.}}}
  \bibinfo{year}{2015}\natexlab{}.
\newblock \bibinfo{title}{Keras}.
\newblock \bibinfo{howpublished}{\url{https://keras.io}}.
\newblock


\bibitem[\protect\citeauthoryear{Christiano, Leike, Brown, Martic, Legg, and
  Amodei}{Christiano et~al\mbox{.}}{2017}]%
        {christiano2017deep}
\bibfield{author}{\bibinfo{person}{Paul~F. Christiano}, \bibinfo{person}{Jan
  Leike}, \bibinfo{person}{Tom~B. Brown}, \bibinfo{person}{Miljan Martic},
  \bibinfo{person}{Shane Legg}, {and} \bibinfo{person}{Dario Amodei}.}
  \bibinfo{year}{2017}\natexlab{}.
\newblock \showarticletitle{Deep reinforcement learning from human
  preferences}. In \bibinfo{booktitle}{\emph{Proceedings of the 31st
  International Conference on Neural Information Processing Systems}}.
  \bibinfo{publisher}{Curran Associates, Inc.}, \bibinfo{pages}{4302--4310}.
\newblock


\bibitem[\protect\citeauthoryear{Custode and Iacca}{Custode and Iacca}{2020}]%
        {custode2020evolutionary}
\bibfield{author}{\bibinfo{person}{Leonardo~L. Custode} {and}
  \bibinfo{person}{Giovanni Iacca}.} \bibinfo{year}{2020}\natexlab{}.
\newblock \showarticletitle{Evolutionary learning of interpretable decision
  trees}.
\newblock \bibinfo{journal}{\emph{arXiv preprint arXiv:2012.07723}}
  (\bibinfo{year}{2020}).
\newblock


\bibitem[\protect\citeauthoryear{De~Freitas, Pappa, da~Silva, Gonc, Moura,
  Veloso, Laender, and de~Carvalho}{De~Freitas et~al\mbox{.}}{2010}]%
        {de2010active}
\bibfield{author}{\bibinfo{person}{Junio De~Freitas},
  \bibinfo{person}{Gisele~L. Pappa}, \bibinfo{person}{Altigran~S. da Silva},
  \bibinfo{person}{Marcos~A. Gonc}, \bibinfo{person}{Edleno Moura},
  \bibinfo{person}{Adriano Veloso}, \bibinfo{person}{Alberto~H.F. Laender},
  {and} \bibinfo{person}{Mois{\'e}s~G. de Carvalho}.}
  \bibinfo{year}{2010}\natexlab{}.
\newblock \showarticletitle{Active learning genetic programming for record
  deduplication}. In \bibinfo{booktitle}{\emph{IEEE Congress on Evolutionary
  Computation}}. IEEE, \bibinfo{publisher}{IEEE}, \bibinfo{pages}{1--8}.
\newblock


\bibitem[\protect\citeauthoryear{Deb, Agrawal, Pratap, and Meyarivan}{Deb
  et~al\mbox{.}}{2000}]%
        {deb2000fast}
\bibfield{author}{\bibinfo{person}{Kalyanmoy Deb}, \bibinfo{person}{Samir
  Agrawal}, \bibinfo{person}{Amrit Pratap}, {and} \bibinfo{person}{Tanaka
  Meyarivan}.} \bibinfo{year}{2000}\natexlab{}.
\newblock \showarticletitle{A fast elitist non-dominated sorting genetic
  algorithm for multi-objective optimization: {NSGA}-{II}}. In
  \bibinfo{booktitle}{\emph{International Conference on Parallel Problem
  Solving from Nature}}. \bibinfo{publisher}{Springer Berlin Heidelberg},
  \bibinfo{pages}{849--858}.
\newblock


\bibitem[\protect\citeauthoryear{Deist, Grewal, Dankers, Alderliesten, and
  Bosman}{Deist et~al\mbox{.}}{2021}]%
        {deist2021multi}
\bibfield{author}{\bibinfo{person}{Timo~M. Deist}, \bibinfo{person}{Monika
  Grewal}, \bibinfo{person}{Frank J. W.~M. Dankers}, \bibinfo{person}{Tanja
  Alderliesten}, {and} \bibinfo{person}{Peter A.~N. Bosman}.}
  \bibinfo{year}{2021}\natexlab{}.
\newblock \showarticletitle{Multi-Objective Learning to Predict {P}areto Fronts
  Using Hypervolume Maximization}.
\newblock \bibinfo{journal}{\emph{arXiv preprint arXiv:2102.04523}}
  (\bibinfo{year}{2021}).
\newblock


\bibitem[\protect\citeauthoryear{Diaconis and Graham}{Diaconis and
  Graham}{1977}]%
        {diaconis1977spearman}
\bibfield{author}{\bibinfo{person}{Persi Diaconis} {and}
  \bibinfo{person}{Ronald~L. Graham}.} \bibinfo{year}{1977}\natexlab{}.
\newblock \showarticletitle{Spearman's footrule as a measure of disarray}.
\newblock \bibinfo{journal}{\emph{Journal of the Royal Statistical Society:
  Series B (Methodological)}} \bibinfo{volume}{39}, \bibinfo{number}{2}
  (\bibinfo{year}{1977}), \bibinfo{pages}{262--268}.
\newblock


\bibitem[\protect\citeauthoryear{Dick, Owen, and Whigham}{Dick
  et~al\mbox{.}}{2020}]%
        {dick2020feature}
\bibfield{author}{\bibinfo{person}{Grant Dick}, \bibinfo{person}{Caitlin~A.
  Owen}, {and} \bibinfo{person}{Peter~A. Whigham}.}
  \bibinfo{year}{2020}\natexlab{}.
\newblock \showarticletitle{Feature Standardisation and Coefficient
  Optimisation for Effective Symbolic Regression}. In
  \bibinfo{booktitle}{\emph{Proceedings of the Genetic and Evolutionary
  Computation Conference}}. \bibinfo{publisher}{Association for Computing
  Machinery}, \bibinfo{pages}{306–314}.
\newblock


\bibitem[\protect\citeauthoryear{Dua and Graff}{Dua and Graff}{2017}]%
        {dua2017uci}
\bibfield{author}{\bibinfo{person}{Dheeru Dua} {and} \bibinfo{person}{Casey
  Graff}.} \bibinfo{year}{2017}\natexlab{}.
\newblock \bibinfo{title}{{UCI} Machine Learning Repository}.
\newblock
\newblock
\urldef\tempurl%
\url{http://archive.ics.uci.edu/ml}
\showURL{%
\tempurl}


\bibitem[\protect\citeauthoryear{Ek{\'a}rt and Nemeth}{Ek{\'a}rt and
  Nemeth}{2001}]%
        {ekart2001selection}
\bibfield{author}{\bibinfo{person}{Anik{\'o} Ek{\'a}rt} {and}
  \bibinfo{person}{Sandor~Z. Nemeth}.} \bibinfo{year}{2001}\natexlab{}.
\newblock \showarticletitle{Selection based on the {P}areto nondomination
  criterion for controlling code growth in genetic programming}.
\newblock \bibinfo{journal}{\emph{Genetic Programming and Evolvable Machines}}
  \bibinfo{volume}{2}, \bibinfo{number}{1} (\bibinfo{year}{2001}),
  \bibinfo{pages}{61--73}.
\newblock


\bibitem[\protect\citeauthoryear{Ferrario, Loi, and Vigan{\`o}}{Ferrario
  et~al\mbox{.}}{2019}]%
        {ferrario2019ai}
\bibfield{author}{\bibinfo{person}{Andrea Ferrario}, \bibinfo{person}{Michele
  Loi}, {and} \bibinfo{person}{Eleonora Vigan{\`o}}.}
  \bibinfo{year}{2019}\natexlab{}.
\newblock \showarticletitle{In {AI} we trust Incrementally: {A} Multi-layer
  model of trust to analyze Human-Artificial intelligence interactions}.
\newblock \bibinfo{journal}{\emph{Philosophy \& Technology}}
  \bibinfo{volume}{33} (\bibinfo{year}{2019}), \bibinfo{pages}{1--17}.
\newblock


\bibitem[\protect\citeauthoryear{Freitas}{Freitas}{2014}]%
        {freitas2014comprehensible}
\bibfield{author}{\bibinfo{person}{Alex~A. Freitas}.}
  \bibinfo{year}{2014}\natexlab{}.
\newblock \showarticletitle{Comprehensible classification models: {A} position
  paper}.
\newblock \bibinfo{journal}{\emph{ACM SIGKDD Explorations Newsletter}}
  \bibinfo{volume}{15}, \bibinfo{number}{1} (\bibinfo{year}{2014}),
  \bibinfo{pages}{1--10}.
\newblock


\bibitem[\protect\citeauthoryear{Gal and Ghahramani}{Gal and
  Ghahramani}{2016}]%
        {gal2016dropout}
\bibfield{author}{\bibinfo{person}{Yarin Gal} {and} \bibinfo{person}{Zoubin
  Ghahramani}.} \bibinfo{year}{2016}\natexlab{}.
\newblock \showarticletitle{Dropout as a {B}ayesian approximation:
  {R}epresenting model uncertainty in deep learning}. In
  \bibinfo{booktitle}{\emph{International Conference on Machine Learning}}.
  \bibinfo{publisher}{JMLR.org}, \bibinfo{pages}{1050--1059}.
\newblock


\bibitem[\protect\citeauthoryear{Guidotti, Monreale, Ruggieri, Turini,
  Giannotti, and Pedreschi}{Guidotti et~al\mbox{.}}{2018}]%
        {guidotti2018survey}
\bibfield{author}{\bibinfo{person}{Riccardo Guidotti}, \bibinfo{person}{Anna
  Monreale}, \bibinfo{person}{Salvatore Ruggieri}, \bibinfo{person}{Franco
  Turini}, \bibinfo{person}{Fosca Giannotti}, {and} \bibinfo{person}{Dino
  Pedreschi}.} \bibinfo{year}{2018}\natexlab{}.
\newblock \showarticletitle{A survey of methods for explaining black box
  models}.
\newblock \bibinfo{journal}{\emph{ACM Computing Surveys (CSUR)}}
  \bibinfo{volume}{51}, \bibinfo{number}{5} (\bibinfo{year}{2018}),
  \bibinfo{pages}{1--42}.
\newblock


\bibitem[\protect\citeauthoryear{Gulrajani, Ahmed, Arjovsky, Dumoulin, and
  Courville}{Gulrajani et~al\mbox{.}}{2017}]%
        {gulrajani2017improved}
\bibfield{author}{\bibinfo{person}{Ishaan Gulrajani}, \bibinfo{person}{Faruk
  Ahmed}, \bibinfo{person}{Martin Arjovsky}, \bibinfo{person}{Vincent
  Dumoulin}, {and} \bibinfo{person}{Aaron Courville}.}
  \bibinfo{year}{2017}\natexlab{}.
\newblock \showarticletitle{Improved training of {W}asserstein {GAN}s}. In
  \bibinfo{booktitle}{\emph{Proceedings of the 31st International Conference on
  Neural Information Processing Systems}} (Long Beach, California, USA)
  \emph{(\bibinfo{series}{NIPS'17})}. \bibinfo{publisher}{Curran Associates
  Inc.}, \bibinfo{pages}{5769–5779}.
\newblock
\showISBNx{9781510860964}


\bibitem[\protect\citeauthoryear{Hatherley}{Hatherley}{2020}]%
        {hatherley2020limits}
\bibfield{author}{\bibinfo{person}{Joshua~James Hatherley}.}
  \bibinfo{year}{2020}\natexlab{}.
\newblock \showarticletitle{Limits of trust in medical {AI}}.
\newblock \bibinfo{journal}{\emph{Journal of Medical Ethics}}
  \bibinfo{volume}{46}, \bibinfo{number}{7} (\bibinfo{year}{2020}),
  \bibinfo{pages}{478--481}.
\newblock


\bibitem[\protect\citeauthoryear{Hein, Udluft, and Runkler}{Hein
  et~al\mbox{.}}{2018}]%
        {hein2018interpretable}
\bibfield{author}{\bibinfo{person}{Daniel Hein}, \bibinfo{person}{Steffen
  Udluft}, {and} \bibinfo{person}{Thomas~A. Runkler}.}
  \bibinfo{year}{2018}\natexlab{}.
\newblock \showarticletitle{Interpretable policies for reinforcement learning
  by genetic programming}.
\newblock \bibinfo{journal}{\emph{Engineering Applications of Artificial
  Intelligence}}  \bibinfo{volume}{76} (\bibinfo{year}{2018}),
  \bibinfo{pages}{158--169}.
\newblock


\bibitem[\protect\citeauthoryear{Huysmans, Dejaeger, Mues, Vanthienen, and
  Baesens}{Huysmans et~al\mbox{.}}{2011}]%
        {huysmans2011empirical}
\bibfield{author}{\bibinfo{person}{Johan Huysmans}, \bibinfo{person}{Karel
  Dejaeger}, \bibinfo{person}{Christophe Mues}, \bibinfo{person}{Jan
  Vanthienen}, {and} \bibinfo{person}{Bart Baesens}.}
  \bibinfo{year}{2011}\natexlab{}.
\newblock \showarticletitle{An empirical evaluation of the comprehensibility of
  decision table, tree and rule based predictive models}.
\newblock \bibinfo{journal}{\emph{Decision Support Systems}}
  \bibinfo{volume}{51}, \bibinfo{number}{1} (\bibinfo{year}{2011}),
  \bibinfo{pages}{141--154}.
\newblock


\bibitem[\protect\citeauthoryear{Isele and Bizer}{Isele and Bizer}{2013}]%
        {isele2013active}
\bibfield{author}{\bibinfo{person}{Robert Isele} {and}
  \bibinfo{person}{Christian Bizer}.} \bibinfo{year}{2013}\natexlab{}.
\newblock \showarticletitle{Active learning of expressive linkage rules using
  genetic programming}.
\newblock \bibinfo{journal}{\emph{Journal of Web Semantics}}
  \bibinfo{volume}{23} (\bibinfo{year}{2013}), \bibinfo{pages}{2--15}.
\newblock


\bibitem[\protect\citeauthoryear{Izza, Ignatiev, and Marques-Silva}{Izza
  et~al\mbox{.}}{2020}]%
        {izza2020explaining}
\bibfield{author}{\bibinfo{person}{Yacine Izza}, \bibinfo{person}{Alexey
  Ignatiev}, {and} \bibinfo{person}{Joao Marques-Silva}.}
  \bibinfo{year}{2020}\natexlab{}.
\newblock \showarticletitle{On explaining decision trees}.
\newblock \bibinfo{journal}{\emph{arXiv preprint arXiv:2010.11034}}
  (\bibinfo{year}{2020}).
\newblock


\bibitem[\protect\citeauthoryear{Jobin, Ienca, and Vayena}{Jobin
  et~al\mbox{.}}{2019}]%
        {jobin2019global}
\bibfield{author}{\bibinfo{person}{Anna Jobin}, \bibinfo{person}{Marcello
  Ienca}, {and} \bibinfo{person}{Effy Vayena}.}
  \bibinfo{year}{2019}\natexlab{}.
\newblock \showarticletitle{The global landscape of {AI} ethics guidelines}.
\newblock \bibinfo{journal}{\emph{Nature Machine Intelligence}}
  \bibinfo{volume}{1}, \bibinfo{number}{9} (\bibinfo{year}{2019}),
  \bibinfo{pages}{389--399}.
\newblock


\bibitem[\protect\citeauthoryear{Keijzer}{Keijzer}{2004}]%
        {keijzer2004scaled}
\bibfield{author}{\bibinfo{person}{Maarten Keijzer}.}
  \bibinfo{year}{2004}\natexlab{}.
\newblock \showarticletitle{Scaled symbolic regression}.
\newblock \bibinfo{journal}{\emph{Genetic Programming and Evolvable Machines}}
  \bibinfo{volume}{5}, \bibinfo{number}{3} (\bibinfo{year}{2004}),
  \bibinfo{pages}{259--269}.
\newblock


\bibitem[\protect\citeauthoryear{Koza}{Koza}{1992}]%
        {koza1992genetic}
\bibfield{author}{\bibinfo{person}{John~R. Koza}.}
  \bibinfo{year}{1992}\natexlab{}.
\newblock \bibinfo{booktitle}{\emph{Genetic programming: {O}n the programming
  of computers by means of natural selection}}. Vol.~\bibinfo{volume}{1}.
\newblock \bibinfo{publisher}{MIT Press}.
\newblock


\bibitem[\protect\citeauthoryear{Lakkaraju, Bach, and Leskovec}{Lakkaraju
  et~al\mbox{.}}{2016}]%
        {lakkaraju2016interpretable}
\bibfield{author}{\bibinfo{person}{Himabindu Lakkaraju},
  \bibinfo{person}{Stephen~H. Bach}, {and} \bibinfo{person}{Jure Leskovec}.}
  \bibinfo{year}{2016}\natexlab{}.
\newblock \showarticletitle{Interpretable decision sets: {A} joint framework
  for description and prediction}. In \bibinfo{booktitle}{\emph{Proceedings of
  the 22nd ACM SIGKDD International Conference on Knowledge Discovery and Data
  Mining}}. \bibinfo{publisher}{Association for Computing Machinery},
  \bibinfo{pages}{1675--1684}.
\newblock


\bibitem[\protect\citeauthoryear{Lensen}{Lensen}{2021}]%
        {lensen2021mining}
\bibfield{author}{\bibinfo{person}{Andrew Lensen}.}
  \bibinfo{year}{2021}\natexlab{}.
\newblock \showarticletitle{Mining Feature Relationships in Data}. In
  \bibinfo{booktitle}{\emph{Genetic Programming: 24th European Conference,
  EuroGP 2021, Held as Part of EvoStar 2021}}, Vol.~\bibinfo{volume}{12691}.
  \bibinfo{publisher}{Springer, Cham}, \bibinfo{pages}{247--262}.
\newblock


\bibitem[\protect\citeauthoryear{Lensen, Xue, and Zhang}{Lensen
  et~al\mbox{.}}{2020}]%
        {lensen2020genetic}
\bibfield{author}{\bibinfo{person}{Andrew Lensen}, \bibinfo{person}{Bing Xue},
  {and} \bibinfo{person}{Mengjie Zhang}.} \bibinfo{year}{2020}\natexlab{}.
\newblock \showarticletitle{Genetic programming for evolving a front of
  interpretable models for data visualization}.
\newblock \bibinfo{journal}{\emph{IEEE Transactions on Cybernetics}}
  (\bibinfo{year}{2020}), \bibinfo{pages}{1--15}.
\newblock


\bibitem[\protect\citeauthoryear{Letham, Rudin, McCormick, and Madigan}{Letham
  et~al\mbox{.}}{2015}]%
        {letham2015interpretable}
\bibfield{author}{\bibinfo{person}{Benjamin Letham}, \bibinfo{person}{Cynthia
  Rudin}, \bibinfo{person}{Tyler~H. McCormick}, {and} \bibinfo{person}{David
  Madigan}.} \bibinfo{year}{2015}\natexlab{}.
\newblock \showarticletitle{Interpretable classifiers using rules and
  {B}ayesian analysis: {B}uilding a better stroke prediction model}.
\newblock \bibinfo{journal}{\emph{Annals of Applied Statistics}}
  \bibinfo{volume}{9}, \bibinfo{number}{3} (\bibinfo{year}{2015}),
  \bibinfo{pages}{1350--1371}.
\newblock


\bibitem[\protect\citeauthoryear{Lipton}{Lipton}{2018}]%
        {lipton2018mythos}
\bibfield{author}{\bibinfo{person}{Zachary~C. Lipton}.}
  \bibinfo{year}{2018}\natexlab{}.
\newblock \showarticletitle{The Mythos of Model Interpretability: {I}n machine
  learning, the concept of interpretability is both important and slippery}.
\newblock \bibinfo{journal}{\emph{Queue}} \bibinfo{volume}{16},
  \bibinfo{number}{3} (\bibinfo{year}{2018}), \bibinfo{pages}{31--57}.
\newblock


\bibitem[\protect\citeauthoryear{Mahoor, Felag, and Bongard}{Mahoor
  et~al\mbox{.}}{2017}]%
        {mahoor2017morphology}
\bibfield{author}{\bibinfo{person}{Zahra Mahoor}, \bibinfo{person}{Jack Felag},
  {and} \bibinfo{person}{Josh Bongard}.} \bibinfo{year}{2017}\natexlab{}.
\newblock \showarticletitle{Morphology dictates a robot's ability to ground
  crowd-proposed language}.
\newblock \bibinfo{journal}{\emph{arXiv preprint arXiv:1712.05881}}
  (\bibinfo{year}{2017}).
\newblock


\bibitem[\protect\citeauthoryear{Medvet, Bartoli, Carminati, and
  Ferrari}{Medvet et~al\mbox{.}}{2015}]%
        {medvet2015evolutionary}
\bibfield{author}{\bibinfo{person}{Eric Medvet}, \bibinfo{person}{Alberto
  Bartoli}, \bibinfo{person}{Barbara Carminati}, {and} \bibinfo{person}{Elena
  Ferrari}.} \bibinfo{year}{2015}\natexlab{}.
\newblock \showarticletitle{Evolutionary inference of attribute-based access
  control policies}. In \bibinfo{booktitle}{\emph{International Conference on
  Evolutionary Multi-Criterion Optimization}}. \bibinfo{publisher}{Springer,
  Cham}, \bibinfo{pages}{351--365}.
\newblock


\bibitem[\protect\citeauthoryear{Molnar}{Molnar}{2020}]%
        {molnar2020interpretable}
\bibfield{author}{\bibinfo{person}{Christoph Molnar}.}
  \bibinfo{year}{2020}\natexlab{}.
\newblock \bibinfo{booktitle}{\emph{Interpretable machine learning --- {A}
  guide for making black box models explainable}}.
\newblock \bibinfo{publisher}{Lulu.com}.
\newblock


\bibitem[\protect\citeauthoryear{Molnar, K{\"o}nig, Herbinger, Freiesleben,
  Dandl, Scholbeck, Casalicchio, Grosse-Wentrup, and Bischl}{Molnar
  et~al\mbox{.}}{2020}]%
        {molnar2020pitfalls}
\bibfield{author}{\bibinfo{person}{Christoph Molnar}, \bibinfo{person}{Gunnar
  K{\"o}nig}, \bibinfo{person}{Julia Herbinger}, \bibinfo{person}{Timo
  Freiesleben}, \bibinfo{person}{Susanne Dandl}, \bibinfo{person}{Christian~A.
  Scholbeck}, \bibinfo{person}{Giuseppe Casalicchio}, \bibinfo{person}{Moritz
  Grosse-Wentrup}, {and} \bibinfo{person}{Bernd Bischl}.}
  \bibinfo{year}{2020}\natexlab{}.
\newblock \showarticletitle{Pitfalls to avoid when interpreting machine
  learning models}.
\newblock \bibinfo{journal}{\emph{arXiv preprint arXiv:2007.04131}}
  (\bibinfo{year}{2020}).
\newblock


\bibitem[\protect\citeauthoryear{Moore, Chua, Berry, and Gair}{Moore
  et~al\mbox{.}}{2016}]%
        {moore2016fast}
\bibfield{author}{\bibinfo{person}{Christopher~J. Moore},
  \bibinfo{person}{Alvin J.~K. Chua}, \bibinfo{person}{Christopher P.~L.
  Berry}, {and} \bibinfo{person}{Jonathan~R. Gair}.}
  \bibinfo{year}{2016}\natexlab{}.
\newblock \showarticletitle{Fast methods for training {G}aussian processes on
  large datasets}.
\newblock \bibinfo{journal}{\emph{Royal Society Open Science}}
  \bibinfo{volume}{3}, \bibinfo{number}{5} (\bibinfo{year}{2016}),
  \bibinfo{pages}{160125}.
\newblock


\bibitem[\protect\citeauthoryear{Mota, Naredo, and Ryan}{Mota
  et~al\mbox{.}}{2021}]%
        {mota2021towards}
\bibfield{author}{\bibinfo{person}{Douglas Mota}, \bibinfo{person}{Enrique
  Naredo}, {and} \bibinfo{person}{Conor Ryan}.}
  \bibinfo{year}{2021}\natexlab{}.
\newblock \showarticletitle{Towards Incorporating Human Knowledge in Fuzzy
  Pattern Tree Evolution}. In \bibinfo{booktitle}{\emph{Genetic Programming:
  24th European Conference, EuroGP 2021, Held as Part of EvoStar 2021}}.
  \bibinfo{publisher}{Springer International Publishing},
  \bibinfo{pages}{66--81}.
\newblock


\bibitem[\protect\citeauthoryear{Nair and Hinton}{Nair and Hinton}{2010}]%
        {nair2010rectified}
\bibfield{author}{\bibinfo{person}{Vinod Nair} {and}
  \bibinfo{person}{Geoffrey~E. Hinton}.} \bibinfo{year}{2010}\natexlab{}.
\newblock \showarticletitle{Rectified linear units improve restricted
  {B}oltzmann machines}. In \bibinfo{booktitle}{\emph{International Conference
  on Machine Learning}}. \bibinfo{publisher}{Omnipress},
  \bibinfo{pages}{807–814}.
\newblock


\bibitem[\protect\citeauthoryear{Poli, Langdon, McPhee, and Koza}{Poli
  et~al\mbox{.}}{2008}]%
        {poli2008field}
\bibfield{author}{\bibinfo{person}{Riccardo Poli}, \bibinfo{person}{William~B
  Langdon}, \bibinfo{person}{Nicholas~F McPhee}, {and} \bibinfo{person}{John~R
  Koza}.} \bibinfo{year}{2008}\natexlab{}.
\newblock \bibinfo{booktitle}{\emph{A field guide to genetic programming}}.
\newblock \bibinfo{publisher}{Lulu.com}.
\newblock


\bibitem[\protect\citeauthoryear{Poursabzi-Sangdeh, Goldstein, Hofman, Vaughan,
  and Wallach}{Poursabzi-Sangdeh et~al\mbox{.}}{2018}]%
        {poursabzi2018manipulating}
\bibfield{author}{\bibinfo{person}{Forough Poursabzi-Sangdeh},
  \bibinfo{person}{Daniel~G. Goldstein}, \bibinfo{person}{Jake~M. Hofman},
  \bibinfo{person}{Jennifer~Wortman Vaughan}, {and} \bibinfo{person}{Hanna
  Wallach}.} \bibinfo{year}{2018}\natexlab{}.
\newblock \showarticletitle{Manipulating and measuring model interpretability}.
\newblock \bibinfo{journal}{\emph{arXiv preprint arXiv:1802.07810}}
  (\bibinfo{year}{2018}).
\newblock


\bibitem[\protect\citeauthoryear{Rakshit, Konar, and Das}{Rakshit
  et~al\mbox{.}}{2017}]%
        {rakshit2017noisy}
\bibfield{author}{\bibinfo{person}{Pratyusha Rakshit}, \bibinfo{person}{Amit
  Konar}, {and} \bibinfo{person}{Swagatam Das}.}
  \bibinfo{year}{2017}\natexlab{}.
\newblock \showarticletitle{Noisy evolutionary optimization algorithms--{A}
  comprehensive survey}.
\newblock \bibinfo{journal}{\emph{Swarm and Evolutionary Computation}}
  \bibinfo{volume}{33} (\bibinfo{year}{2017}), \bibinfo{pages}{18--45}.
\newblock


\bibitem[\protect\citeauthoryear{Rudin}{Rudin}{2019}]%
        {rudin2019stop}
\bibfield{author}{\bibinfo{person}{Cynthia Rudin}.}
  \bibinfo{year}{2019}\natexlab{}.
\newblock \showarticletitle{Stop explaining black box machine learning models
  for high stakes decisions and use interpretable models instead}.
\newblock \bibinfo{journal}{\emph{Nature Machine Intelligence}}
  \bibinfo{volume}{1}, \bibinfo{number}{5} (\bibinfo{year}{2019}),
  \bibinfo{pages}{206--215}.
\newblock


\bibitem[\protect\citeauthoryear{Secretan, Beato, D'Ambrosio, Rodriguez,
  Campbell, Folsom-Kovarik, and Stanley}{Secretan et~al\mbox{.}}{2011}]%
        {secretan2011picbreeder}
\bibfield{author}{\bibinfo{person}{Jimmy Secretan}, \bibinfo{person}{Nicholas
  Beato}, \bibinfo{person}{David~B. D'Ambrosio}, \bibinfo{person}{Adelein
  Rodriguez}, \bibinfo{person}{Adam Campbell}, \bibinfo{person}{Jeremiah~T.
  Folsom-Kovarik}, {and} \bibinfo{person}{Kenneth~O. Stanley}.}
  \bibinfo{year}{2011}\natexlab{}.
\newblock \showarticletitle{Picbreeder: {A} case study in collaborative
  evolutionary exploration of design space}.
\newblock \bibinfo{journal}{\emph{Evolutionary Computation}}
  \bibinfo{volume}{19}, \bibinfo{number}{3} (\bibinfo{year}{2011}),
  \bibinfo{pages}{373--403}.
\newblock


\bibitem[\protect\citeauthoryear{Settles}{Settles}{2009}]%
        {settles2009active}
\bibfield{author}{\bibinfo{person}{Burr Settles}.}
  \bibinfo{year}{2009}\natexlab{}.
\newblock \bibinfo{booktitle}{\emph{Active learning literature survey}}.
\newblock \bibinfo{type}{{T}echnical {R}eport}.
  \bibinfo{institution}{University of Wisconsin-Madison Department of Computer
  Sciences}.
\newblock


\bibitem[\protect\citeauthoryear{Setzu, Guidotti, Monreale, Turini, Pedreschi,
  and Giannotti}{Setzu et~al\mbox{.}}{2021}]%
        {setzu2021glocalx}
\bibfield{author}{\bibinfo{person}{Mattia Setzu}, \bibinfo{person}{Riccardo
  Guidotti}, \bibinfo{person}{Anna Monreale}, \bibinfo{person}{Franco Turini},
  \bibinfo{person}{Dino Pedreschi}, {and} \bibinfo{person}{Fosca Giannotti}.}
  \bibinfo{year}{2021}\natexlab{}.
\newblock \showarticletitle{{GL}ocal{X}---{F}rom Local to Global Explanations
  of Black Box {AI} Models}.
\newblock \bibinfo{journal}{\emph{Artificial Intelligence}}
  \bibinfo{volume}{294} (\bibinfo{year}{2021}), \bibinfo{pages}{103457}.
\newblock


\bibitem[\protect\citeauthoryear{Smits and Kotanchek}{Smits and
  Kotanchek}{2005}]%
        {smits2005pareto}
\bibfield{author}{\bibinfo{person}{Guido~F. Smits} {and} \bibinfo{person}{Mark
  Kotanchek}.} \bibinfo{year}{2005}\natexlab{}.
\newblock \showarticletitle{{P}areto-front exploitation in symbolic
  regression}.
\newblock In \bibinfo{booktitle}{\emph{Genetic Programming Theory and Practice
  II}}. \bibinfo{publisher}{Springer}, \bibinfo{pages}{283--299}.
\newblock


\bibitem[\protect\citeauthoryear{Spearman}{Spearman}{1906}]%
        {spearman1906footrule}
\bibfield{author}{\bibinfo{person}{Charles Spearman}.}
  \bibinfo{year}{1906}\natexlab{}.
\newblock \showarticletitle{Footrule for measuring correlation}.
\newblock \bibinfo{journal}{\emph{British Journal of Psychology}}
  \bibinfo{volume}{2}, \bibinfo{number}{1} (\bibinfo{year}{1906}),
  \bibinfo{pages}{89}.
\newblock


\bibitem[\protect\citeauthoryear{Srivastava, Hinton, Krizhevsky, Sutskever, and
  Salakhutdinov}{Srivastava et~al\mbox{.}}{2014}]%
        {srivastava2014dropout}
\bibfield{author}{\bibinfo{person}{Nitish Srivastava},
  \bibinfo{person}{Geoffrey Hinton}, \bibinfo{person}{Alex Krizhevsky},
  \bibinfo{person}{Ilya Sutskever}, {and} \bibinfo{person}{Ruslan
  Salakhutdinov}.} \bibinfo{year}{2014}\natexlab{}.
\newblock \showarticletitle{Dropout: {A} Simple Way to Prevent Neural Networks
  from Overfitting}.
\newblock \bibinfo{journal}{\emph{Journal of Machine Learning Research}}
  \bibinfo{volume}{15}, \bibinfo{number}{1} (\bibinfo{year}{2014}),
  \bibinfo{pages}{1929–1958}.
\newblock


\bibitem[\protect\citeauthoryear{Su, Wei, Varshney, and Malioutov}{Su
  et~al\mbox{.}}{2015}]%
        {su2015interpretable}
\bibfield{author}{\bibinfo{person}{Guolong Su}, \bibinfo{person}{Dennis Wei},
  \bibinfo{person}{Kush~R. Varshney}, {and} \bibinfo{person}{Dmitry~M.
  Malioutov}.} \bibinfo{year}{2015}\natexlab{}.
\newblock \showarticletitle{Interpretable two-level boolean rule learning for
  classification}.
\newblock \bibinfo{journal}{\emph{arXiv preprint arXiv:1511.07361}}
  (\bibinfo{year}{2015}).
\newblock


\bibitem[\protect\citeauthoryear{Tibshirani}{Tibshirani}{1996}]%
        {tibshirani1996regression}
\bibfield{author}{\bibinfo{person}{Robert Tibshirani}.}
  \bibinfo{year}{1996}\natexlab{}.
\newblock \showarticletitle{Regression shrinkage and selection via the
  {LASSO}}.
\newblock \bibinfo{journal}{\emph{Journal of the Royal Statistical Society:
  Series B (Methodological)}} \bibinfo{volume}{58}, \bibinfo{number}{1}
  (\bibinfo{year}{1996}), \bibinfo{pages}{267--288}.
\newblock


\bibitem[\protect\citeauthoryear{Ustun and Rudin}{Ustun and Rudin}{2016}]%
        {ustun2016supersparse}
\bibfield{author}{\bibinfo{person}{Berk Ustun} {and} \bibinfo{person}{Cynthia
  Rudin}.} \bibinfo{year}{2016}\natexlab{}.
\newblock \showarticletitle{Supersparse linear integer models for optimized
  medical scoring systems}.
\newblock \bibinfo{journal}{\emph{Machine Learning}} \bibinfo{volume}{102},
  \bibinfo{number}{3} (\bibinfo{year}{2016}), \bibinfo{pages}{349--391}.
\newblock


\bibitem[\protect\citeauthoryear{Virgolin, Alderliesten, and Bosman}{Virgolin
  et~al\mbox{.}}{2019}]%
        {virgolin2019linear}
\bibfield{author}{\bibinfo{person}{Marco Virgolin}, \bibinfo{person}{Tanja
  Alderliesten}, {and} \bibinfo{person}{Peter A.~N. Bosman}.}
  \bibinfo{year}{2019}\natexlab{}.
\newblock \showarticletitle{Linear scaling with and within semantic
  backpropagation-based genetic programming for symbolic regression}. In
  \bibinfo{booktitle}{\emph{Proceedings of the Genetic and Evolutionary
  Computation Conference}}. \bibinfo{publisher}{Association for Computing
  Machinery}, \bibinfo{pages}{1084--1092}.
\newblock


\bibitem[\protect\citeauthoryear{Virgolin, Alderliesten, and Bosman}{Virgolin
  et~al\mbox{.}}{2020a}]%
        {virgolin2020explaining}
\bibfield{author}{\bibinfo{person}{Marco Virgolin}, \bibinfo{person}{Tanja
  Alderliesten}, {and} \bibinfo{person}{Peter A.~N. Bosman}.}
  \bibinfo{year}{2020}\natexlab{a}.
\newblock \showarticletitle{On explaining machine learning models by evolving
  crucial and compact features}.
\newblock \bibinfo{journal}{\emph{Swarm and Evolutionary Computation}}
  \bibinfo{volume}{53} (\bibinfo{year}{2020}), \bibinfo{pages}{100640}.
\newblock


\bibitem[\protect\citeauthoryear{Virgolin, Alderliesten, Witteveen, and
  Bosman}{Virgolin et~al\mbox{.}}{2020b}]%
        {virgolin2020improving}
\bibfield{author}{\bibinfo{person}{Marco Virgolin}, \bibinfo{person}{Tanja
  Alderliesten}, \bibinfo{person}{Cees Witteveen}, {and} \bibinfo{person}{Peter
  A.~N. Bosman}.} \bibinfo{year}{2020}\natexlab{b}.
\newblock \showarticletitle{Improving model-based genetic programming for
  symbolic regression of small expressions}.
\newblock \bibinfo{journal}{\emph{Evolutionary computation}}
  (\bibinfo{year}{2020}), \bibinfo{pages}{tba}.
\newblock


\bibitem[\protect\citeauthoryear{Virgolin, De~Lorenzo, Medvet, and
  Randone}{Virgolin et~al\mbox{.}}{2020c}]%
        {virgolin2020learning}
\bibfield{author}{\bibinfo{person}{Marco Virgolin}, \bibinfo{person}{Andrea
  De~Lorenzo}, \bibinfo{person}{Eric Medvet}, {and} \bibinfo{person}{Francesca
  Randone}.} \bibinfo{year}{2020}\natexlab{c}.
\newblock \showarticletitle{Learning a formula of interpretability to learn
  interpretable formulas}. In \bibinfo{booktitle}{\emph{International
  Conference on Parallel Problem Solving from Nature}}.
  \bibinfo{publisher}{Springer, Cham}, \bibinfo{pages}{79--93}.
\newblock


\bibitem[\protect\citeauthoryear{Vladislavleva, Smits, and
  Den~Hertog}{Vladislavleva et~al\mbox{.}}{2008}]%
        {vladislavleva2008order}
\bibfield{author}{\bibinfo{person}{Ekaterina~J. Vladislavleva},
  \bibinfo{person}{Guido~F. Smits}, {and} \bibinfo{person}{Dick Den~Hertog}.}
  \bibinfo{year}{2008}\natexlab{}.
\newblock \showarticletitle{Order of nonlinearity as a complexity measure for
  models generated by symbolic regression via {P}areto genetic programming}.
\newblock \bibinfo{journal}{\emph{IEEE Transactions on Evolutionary
  Computation}} \bibinfo{volume}{13}, \bibinfo{number}{2}
  (\bibinfo{year}{2008}), \bibinfo{pages}{333--349}.
\newblock


\bibitem[\protect\citeauthoryear{Wang and Rudin}{Wang and Rudin}{2015}]%
        {wang2015falling}
\bibfield{author}{\bibinfo{person}{Fulton Wang} {and} \bibinfo{person}{Cynthia
  Rudin}.} \bibinfo{year}{2015}\natexlab{}.
\newblock \showarticletitle{Falling rule lists}. In
  \bibinfo{booktitle}{\emph{Artificial Intelligence and Statistics}}. PMLR,
  \bibinfo{pages}{1013--1022}.
\newblock


\bibitem[\protect\citeauthoryear{Wang, Rudin, Doshi-Velez, Liu, Klampfl, and
  MacNeille}{Wang et~al\mbox{.}}{2017}]%
        {wang2017bayesian}
\bibfield{author}{\bibinfo{person}{Tong Wang}, \bibinfo{person}{Cynthia Rudin},
  \bibinfo{person}{Finale Doshi-Velez}, \bibinfo{person}{Yimin Liu},
  \bibinfo{person}{Erica Klampfl}, {and} \bibinfo{person}{Perry MacNeille}.}
  \bibinfo{year}{2017}\natexlab{}.
\newblock \showarticletitle{A {B}ayesian framework for learning rule sets for
  interpretable classification}.
\newblock \bibinfo{journal}{\emph{The Journal of Machine Learning Research}}
  \bibinfo{volume}{18}, \bibinfo{number}{1} (\bibinfo{year}{2017}),
  \bibinfo{pages}{2357--2393}.
\newblock


\bibitem[\protect\citeauthoryear{Yang and Loog}{Yang and Loog}{2016}]%
        {yang2016active}
\bibfield{author}{\bibinfo{person}{Yazhou Yang} {and} \bibinfo{person}{Marco
  Loog}.} \bibinfo{year}{2016}\natexlab{}.
\newblock \showarticletitle{Active learning using uncertainty information}. In
  \bibinfo{booktitle}{\emph{23rd International Conference on Pattern
  Recognition (ICPR)}}. \bibinfo{publisher}{IEEE}, \bibinfo{pages}{2646--2651}.
\newblock


\bibitem[\protect\citeauthoryear{Zou and Hastie}{Zou and Hastie}{2005}]%
        {zou2005regularization}
\bibfield{author}{\bibinfo{person}{Hui Zou} {and} \bibinfo{person}{Trevor
  Hastie}.} \bibinfo{year}{2005}\natexlab{}.
\newblock \showarticletitle{Regularization and variable selection via the
  elastic net}.
\newblock \bibinfo{journal}{\emph{Journal of the Royal Statistical Society:
  Series B (Methodological)}} \bibinfo{volume}{67}, \bibinfo{number}{2}
  (\bibinfo{year}{2005}), \bibinfo{pages}{301--320}.
\newblock


\end{thebibliography}

\end{document}